\documentclass{article}

\usepackage[final,nonatbib]{neurips_2022}

\usepackage[sort&compress,round]{natbib}
\usepackage{url}            % simple URL typesetting
\usepackage{hyperref}       % hyperlinks
\usepackage[utf8]{inputenc} % allow utf-8 input
\usepackage{xcolor}         % hyperlinks
\usepackage{amsfonts}       % blackboard math symbols
\usepackage{nicefrac}       % compact symbols for 1/2, etc.
\usepackage{setspace}
\usepackage{wrapfig}
\usepackage[inline]{enumitem}
\usepackage{caption}     
\usepackage{subcaption}     
\usepackage{float}     
\usepackage{pifont}     
\usepackage{pgfplots}     

\usepackage{amsmath,amssymb,amsthm}
\usepackage{mathtools}

\ifxetex
  \usepackage{fontspec}
  \usepackage{microtype}
  \usepackage{unicode-math}
  \newcommand{\mathmat}[1]{\symbfup{#1}}
  \newcommand{\mathvec}[1]{\symbfup{#1}}
  \newcommand{\mathvecgreek}[1]{\symbfup{#1}}
  \newcommand{\mathmatgreek}[1]{\symbfup{#1}}

  \newcommand{\mathrvvec}[1]{\mathbfsf{#1}}
  
  \newcommand{\mathrvvecgreek}[1]{\mathbfsf{#1}}
\else
  \usepackage[T1]{fontenc}
  \usepackage[notext,lcgreekalpha]{stix2}
  \usepackage{newtxtext}
  \usepackage{textcomp}

  \newcommand{\mathmat}[1]{\mathbf{#1}}
  \newcommand{\mathvec}[1]{\mathbf{#1}}
  \newcommand{\mathvecgreek}[1]{\mathbf{#1}}
  \newcommand{\mathmatgreek}[1]{\mathbf{#1}}

  \newcommand{\mathrvvec}[1]{\mathbfsf{#1}}
  
  \newcommand{\mathrvvecgreek}[1]{\mathbfsf{#1}}
\fi

\usepackage{booktabs,threeparttable,arydshln}
\usepackage{multirow}
\usepackage{nicematrix}

\usepackage{dsfont}
\usepackage[algo2e, ruled]{algorithm2e}

\usepackage{pgfplots}
\pgfkeys{/pgfplots/tuftelike/.style={
  semithick,
  tick style={major tick length=4pt,semithick,black},
  separate axis lines,
  axis x line*=bottom,
  axis x line shift=2pt,
  xlabel shift=-2pt,
  axis y line*=left,
  axis y line shift=2pt,
  ylabel shift=-2pt}}
  
\usepackage{proof-at-the-end}
\declaretheoremstyle[
    spaceabove=0\topsep,
    spacebelow=0\topsep,
    name={Proposition},
]{propsty}

\declaretheoremstyle[
    spaceabove=0\topsep,
    spacebelow=0\topsep,
    name={Theorem},
]{thmsty}

\newtheorem{assumption}{\textbf{Assumption}}

\usepackage{mdframed}
\newmdtheoremenv{framedtheorem}{\textbf{Theorem}}
\newmdtheoremenv{framedproposition}{\textbf{Proposition}}
\newmdtheoremenv{framedlemma}{\textbf{Lemma}}

\usepackage{url}
\usepackage{hyperref}
\usepackage[noabbrev, capitalise, nameinlink]{cleveref}

\crefname{assumption}{Assumption}{Assumption}
\crefname{condition}{Condition}{Condition}

\crefname{framedtheorem}{Theorem}{Theorems}
\crefname{framedproposition}{Proposition}{Propositions}
\crefname{framedlemma}{Lemma}{Lemmas}

\makeatletter
\def\adl@drawiv#1#2#3{%
        \hskip.5\tabcolsep
        \xleaders#3{#2.5\@tempdimb #1{1}#2.5\@tempdimb}%
                #2\z@ plus1fil minus1fil\relax
        \hskip.5\tabcolsep}
\newcommand{\cdashlinelr}[1]{%
  \noalign{\vskip\aboverulesep
           \global\let\@dashdrawstore\adl@draw
           \global\let\adl@draw\adl@drawiv}
  \cdashline{#1}
  \noalign{\global\let\adl@draw\@dashdrawstore
           \vskip\belowrulesep}}
\makeatother

\pgfkeys{/prAtEnd/global custom defaults/.style={
    end,
    restate,
    text proof={Proof},
    text link={\textit{Proof.} See the \hyperref[proof:prAtEnd\pratendcountercurrent]{\textit{full proof}} in page~\pageref{proof:prAtEnd\pratendcountercurrent}.}
  }
}

\newcommand*\xbar[1]{%
  \hbox{%
    \vbox{%
      \hrule height 0.6pt % The actual bar
      \kern0.33ex%         % Distance between bar and symbol
      \hbox{%
        \kern-0.1em%      % Shortening on the left side
        \ensuremath{#1}%
        \kern-0.1em%      % Shortening on the right side
      }%
    }%
  }%
} 

\newcommand{\E}[1]{\mathbb{E}\left[\,#1\,\right]}
\newcommand{\Esub}[2]{\mathbb{E}_{#1}\left[\,#2\,\right]}
\newcommand{\V}[1]{\mathbb{V}\left[\,#1\,\right]}
\newcommand{\Vsub}[2]{\mathbb{V}_{#1}\left[\,#2\,\right]}
\newcommand{\Cov}[1]{\mathrm{Cov}\left(\,#1\,\right)}

\newcommand{\DKL}[2]{d_{\mathrm{KL}}(#1\parallel#2)}
\newcommand{\DChi}[2]{d_{\chi^2}(#1\parallel#2)}
\newcommand{\norm}[1]{{\left\lVert\,#1\,\right\rVert}}
\newcommand{\abs}[1]{{\left|\,#1\,\right|}}
\newcommand{\DTV}[2]{{d_{\mathrm{TV}}\left(#1, #2\right)}}

\newcommand{\vg}{\mathvec{g}}

\newcommand{\vs}{\mathvec{s}}

\newcommand{\vx}{\mathvec{x}}

\newcommand{\vz}{\mathvec{z}}

\newcommand{\veta}{\mathvecgreek{\eta}}
\newcommand{\vmu}{\mathvecgreek{\mu}}

\newcommand{\vlambda}{\mathvecgreek{\lambda}}

\newcommand{\rvvg}{\mathrvvec{g}}

\newcommand{\rvvz}{\mathrvvec{z}}
\newcommand{\rvvlambda}{\mathrvvecgreek{\lambda}}
\newcommand{\rvveta}{\mathrvvecgreek{\eta}}

\newcommand{\mI}{\mathmat{I}}

\newcommand{\mW}{\mathmat{W}}

\newcommand{\mSigma}{\mathmatgreek{\Sigma}}

\allowdisplaybreaks

\usepackage{tikz}
\usepackage{pgf}
\usepackage{pgfplots}
\usepackage{pgfplotstable}
\usetikzlibrary{patterns}
\usetikzlibrary{fit, calc}
\usetikzlibrary{pgfplots.groupplots}
\usepgfplotslibrary{fillbetween}
\usepgfplotslibrary{groupplots}
\usepgfplotslibrary{statistics}
\usepgfplotslibrary{external} 

\definecolor{darkgrey}{HTML}{3E3E3E}
\definecolor{linkcolor}{HTML}{005D5D}
\definecolor{citecolor}{HTML}{00539A}

\hypersetup{colorlinks={true},linkcolor=linkcolor,citecolor=citecolor}

\usepackage{pifont}
\usepackage{xcolor}

\tikzset{external/system call={xelatex -shell-escape -halt-on-error -interaction=batchmode -jobname "\image" "\texsource"}}
\tikzexternaldisable

\title{\textsc{Markov Chain Score Ascent}: \\ A Unifying Framework of \\ Variational Inference with Markovian Gradients}

\author{%
  Kyurae Kim\thanks{K. Kim is currently with the University of Pennsylvania.} \\
  Sogang University\\
  \texttt{msca8h@sogang.ac.kr} \\
  \And
  Jisu Oh\thanks{J. Oh is currently with North Carolina State University.} \\
  Sogang University \\
  \texttt{jisuoh@sogang.ac.kr} \\
  \AND
  Jacob R. Gardner  \\
  University of Pennsylvania \\
  \texttt{jacobrg@seas.upenn.edu} \\
  \And
  Adji Bousso Dieng \\
  Princeton University \\
  \texttt{adji@princeton.edu} \\
  \And
  Hongseok Kim\thanks{Corresponding author.} \\
  Sogang University \\
  \texttt{hongseok@sogang.ac.kr} \\
}

\begin{document}

\maketitle

\begin{abstract}
  
Minimizing the inclusive Kullback-Leibler (KL) divergence with stochastic gradient descent (SGD) is challenging since its gradient is defined as an integral over the posterior.
Recently, multiple methods have been proposed to run SGD with \textit{biased} gradient estimates obtained from a Markov chain.
This paper provides the first non-asymptotic convergence analysis of these methods by establishing their mixing rate and gradient variance.
To do this, we demonstrate that these methods--which we collectively refer to as Markov chain score ascent (MCSA) methods--can be cast as special cases of the Markov chain gradient descent framework.
Furthermore, by leveraging this new understanding, we develop a novel MCSA scheme, \textit{parallel} MCSA (pMCSA), that achieves a tighter bound on the gradient variance.
We demonstrate that this improved theoretical result translates to superior empirical performance.

% These methods, we call Markov chain score ascent (MCSA), obtain noisy estimates of the gradient by running Markov chains in conjunction with SGD iterations.
% This paper shows that these MCSA methods are a special case of the Markov chain gradient descent (MCGD) framework.
% Based on the non-asymptotic convergence results of MCGD, we analyze the practical performance of previously developed MCSA algorithms.
% Furthermore, we propose a novel MCSA scheme, parallel MCSA (pMCSA), that achieves a tighter bound on the gradient variance.
% We evaluate pMCSA against previously developed MCSA methods on general Bayesian inference benchmarks and demonstrate its performance.

%Our experiments show that, when using our proposed scheme, inclusive KL divergence minimization is competitive against evidence lower bound minimization.
%Our results motivate the use of the inclusive KL divergence for VI.

%%% Local Variables:
%%% TeX-master: "master"
%%% End:

\end{abstract}

\section{Introduction}\label{section:intro}
Bayesian inference aims to analyze the posterior distribution of an unknown latent variable \(\vz\) from which data \(\vx\) is observed.
By assuming a model \(p\,(\vx\,|\,\vz)\), the posterior \(\pi\,(\vz)\) is given by Bayes' rule such that \(\pi\,(\vz) \propto {p\,(\vx\,|\,\vz)\,p\,(\vz)}\) where \(p\,(\vz)\) represents our prior belief on \(\vz\).
Instead of working directly with \(\pi\), variational inference (VI,~\citealt{blei_variational_2017}) seeks a \textit{variational approximation} \(q\,(\vz; \vlambda) \in \mathcal{Q}\), where \(\mathcal{Q}\) is a variational family and \(\vlambda\) are the variational parameters, that is the most similar to \(\pi\) according to a discrepancy measure \(d\,(\pi,\, q\,(\cdot; \vlambda))\).

%% Naturally, choosing an appropriate discrepancy measure is key, which has led to a quest for suitable divergence measures~\citep{ NIPS2016_7750ca35, NIPS2017_35464c84, NEURIPS2018_1cd138d0, NEURIPS2020_c928d86f, regli_alphabeta_2018, pmlr-v48-hernandez-lobatob16, NIPS2016_7750ca35 }.
\looseness=-1
The apparent importance of choosing the right discrepancy measure has led to a quest spanning a decade~\citep{NIPS2017_35464c84, NEURIPS2018_1cd138d0, NEURIPS2020_c928d86f, regli_alphabeta_2018, pmlr-v48-hernandez-lobatob16,NIPS2016_7750ca35,pmlr-v37-salimans15,pmlr-v97-ruiz19a,NEURIPS2021_05f971b5,NEURIPS2021_a1a609f1, bamler_perturbative_2017}.
So far, the exclusive (or reverse, backward)  Kullback-Leibler (KL) divergence \(\DKL{q\left(\cdot; \vlambda\right)}{\pi}\) has seen ``exclusive'' use, partly because it is defined as an integral over \(q\,(\vz; \vlambda)\), which can be approximated efficiently.
In contrast, the \textit{inclusive} (or forward) KL is defined as an integral over \(\pi\) as
{%\small
\vspace{-0.05in}
\begin{align*}
  \DKL{\pi}{q\left(\cdot; \vlambda\right)}
  = \int \pi\,(\vz) \log \frac{\pi\left(\vz\right)}{q\,(\vz; \vlambda)} \,d\vz
  = \Esub{\rvvz \sim \pi\left(\cdot\right)}{\log \frac{\pi\left(\rvvz\right)}{\,q\,(\rvvz; \vlambda)} }. %\label{eq:klpq}
\end{align*}
%\vspace{-0.05in}
}%
Since our goal is to approximate \(\pi\) with \(q\left(\cdot;\vlambda\right)\) but the inclusive KL involves an integral over \(\pi\), we end up facing a chicken-and-egg problem.
Despite this challenge, the inclusive KL has consistently drawn attention due to its statistical properties, such as better uncertainty estimates due to its mass covering property~\citep{minka2005divergence, mackay_local_2001, trippe_overpruning_2017}.

Recently,~\citet{NEURIPS2020_b2070693,pmlr-v124-ou20a} have respectively proposed Markovian score climbing (MSC) and joint stochastic approximation (JSA).
These methods minimize the inclusive KL using stochastic gradient descent (SGD,~\citealt{robbins_stochastic_1951}), where the gradients are estimated using a Markov chain.
The Markov chain kernel \(K_{\vlambda_t}\left(\vz_t, \cdot\right)\) is \(\pi\)-invariant~\citep{robert_monte_2004} and is chosen such that it directly takes advantage of the current variational approximation \(q\left(\cdot; \vlambda_t\right)\).
Thus, the quality of the gradients improves over time as the KL divergence decreases.
Still, the gradients are non-asymptotically biased and Markovian across adjacent iterations, which sharply contrasts MSC and JSA from classical black-box VI~\citep{pmlr-v33-ranganath14, JMLR:v18:16-107}, where the gradients are unbiased and independent.
While~\citet{NEURIPS2020_b2070693} have shown the convergence of MSC through the work of~\citet{gu_stochastic_1998}, this result is only asymptotic and does not provide practical insight into the performance of MSC.

In this paper, we address these theoretical gaps by casting MSC and JSA into a general framework we call Markov chain score ascent (MCSA), which we show is a special case of Markov chain gradient descent (MCGD,~\citealt{duchi_ergodic_2012}).
This enables the application of the non-asymptotic convergence results of MCGD~\citep{duchi_ergodic_2012, NEURIPS2018_1371bcce, pmlr-v99-karimi19a, doan_finitetime_2020, doan_convergence_2020, Xiong_Xu_Liang_Zhang_2021, debavelaere_convergence_2021}.
For MCGD methods, the fundamental properties affecting the convergence rate are the ergodic convergence rate (\(\rho\)) of the MCMC kernel and the gradient variance (\(G\)).
We analyze \(\rho\) and \(G\) of MSC and JSA, enabling their practical comparison given a fixed computational budget (\(N\)).
Furthermore, based on the recent insight that the mixing rate does not affect the convergence rate of MCGD~\citep{doan_convergence_2020,doan_finitetime_2020}, we propose a novel scheme, parallel MCSA (pMCSA), which achieves lower variance by trading off the mixing rate.
We verify our theoretical analysis through numerical simulations and compare MSC, JSA, and pMCSA on general Bayesian inference problems.
Our experiments show that our proposed method outperforms previous MCSA approaches.
%Furthermore, MSCA is competitive against evidence lower-bound (ELBO) maximization within our experiments.
%We further discuss this result in relation with the conclusions of~\citet{dhaka_challenges_2021} in~\cref{section:discussion}.

\vspace{-1.7ex}
\paragraph{Contribution Summary}
%\jrg{Item one should be something along the lines of we provide the first non-asymptotic analysis of MSC and JSA. Item two should be we accomplish this by showing it's a special case (basically point 1 below). I've sketched this out -- maybe we add back in the appropriate sec/thm references?}
%\vspace{-0.2in}
%% \begin{enumerate*}[label=\textbf{(\roman*)}]
\setlength\itemsep{0.01in}
\begin{enumerate}[leftmargin=7mm,listparindent=\parindent]
    \vspace{-1.5ex}
    \item[\ding{182}] We provide the first non-asymptotic theoretical analysis of two recently proposed inclusive KL minimization methods~(\textbf{\cref{section:comparison}}), MSC~(\textbf{\cref{thm:msc,thm:mscrb}}) and JSA~(\textbf{\cref{thm:jsa}}).
    \item[\ding{183}] To do this, we show that both methods can be viewed as what we call ``Markov chain score ascent'' (MCSA) methods~(\textbf{\cref{section:mcsa}}), which are a special case of MCGD~(\textbf{\cref{thm:product_kernel}}).
    \item[\ding{184}] In light of this, we develop a novel MCSA method which we call parallel MCSA (pMCSA, \textbf{\cref{section:pmcsa}}) that achieves lower gradient variance~(\textbf{\cref{thm:pmcsa}}).
    \item[\ding{185}] We demonstrate that the improved theoretical performance of pMCSA translates to superior empirical performance across a variety of Bayesian inference tasks~(\textbf{\cref{section:eval}}).
\end{enumerate}

%% \vspace{-1.5ex}%
%% {%
%% \paragraph{Notation}%
%% \textcolor{blue}{%
%% \(\mathcal{F}_t\) is the \(\sigma\)-field formed by the iterates \(\veta_t\) up to the \(t\)th MCGD iteration, \(\norm{\vx}_* \) is the dual norm of \(\vx\) such that, \(\norm{\vx}_* = \sup_{\norm{\vz} \leq 1} \iprod{\vx}{\vz}\), \(d_{\text{TV}}\left(\pi, q\right)\) is the total-variation (TV) distance between \(\pi\) and \(q\), and \(\DChi{\pi}{q}\) is the \(\chi^2\) divergence from \(\pi\) to \(q\).
%% }}

%%% Local Variables:
%%% TeX-master: "master"
%%% End:

\vspace{-2ex}
\section{Background}
\vspace{-2ex}
\subsection{Inclusive Kullback-Leibler Minimization with Stochastic Gradients}\label{section:ivi_previous}
\vspace{-0.05in}
\paragraph{VI with SGD}
The goal of VI is to find the optimal variational parameters \(\vlambda\) identifying \(q\left(\cdot; \vlambda\right) \in \mathcal{Q}\) that minimizes some discrepancy measure \(D\left(\pi, q\left(\cdot; \vlambda\right)\right)\).
A typical way to perform VI is to use stochastic gradient descent (SGD,~\citealt{robbins_stochastic_1951}), provided that the optimization objective provides \textit{unbiased} gradient estimates \(\vg\,(\vlambda)\) such that we can repeat the update
%\vspace{-0.05in}
{%\small
\begin{align*}
  \vlambda_{t} = \vlambda_{t-1} - \gamma_{t} \, \vg\left(\vlambda_{t-1}\right),
\end{align*}
}%
where \(\gamma_1, \ldots, \gamma_T\) is a stepsize schedule.

\vspace{-2ex}
\paragraph{Inclusive KL Minimization with SGD}
For inclusive KL minimization, \(\vg\) should be set as
{%\small
\begin{align*}
  \vg\left(\vlambda\right)
  = \nabla_{\vlambda} \DKL{\pi}{q\left(\cdot; \vlambda\right)}
  = \nabla_{\vlambda} \mathbb{H}\left[\pi, q\left(\cdot; \vlambda\right)\right]
  = -\Esub{\rvvz \sim \pi\left(\cdot\right)}{ \vs\,(\vlambda; \rvvz) },
\end{align*}
}%
%\vspace{-0.05in}
%
\looseness=-1
where \(\mathbb{H}\left[\pi, q\left(\cdot; \vlambda\right)\right]\) is the cross-entropy between \(\pi\) and \(q\left(\cdot; \vlambda\right)\), which shows the connection with cross-entropy methods~\citep{deboer_tutorial_2005}, and \(\vs\,(\vlambda;\vz) = \nabla_{\vlambda} \log q\,(\vz; \vlambda)\) is known as the \textit{score gradient}.
Since inclusive KL minimization with SGD is equivalent to ascending towards the direction of the score, \citet{NEURIPS2020_b2070693} coined the term score climbing.
To better conform with the optimization literature, we instead call this approach \textit{score ascent} as in gradient ascent.

%\vspace{-0.05in}
%% \subsection{Estimating the Score with Importance Sampling}\label{section:is}
%% When it is easy to sample from the variational approximation \(q_{\lambda}(\vz)\), one can use importance sampling (IS, \citealt{robert_monte_2004}) for estimating the score following
%% %\vspace{-0.05in}
%% {%\small
%% \begin{align*}
%%   \Esub{\pi}{ \vs\,(\vlambda;\rvvz) } 
%%   \propto Z\,\Esub{\pi}{ \vs\,(\vlambda; \rvvz) } %\label{eq:is_est} \\
%%   \approx \frac{1}{N} \sum^{N}_{i=1} \widetilde{w}\,(\vz^{(i)}) \, \vs\,(\vlambda; \vz^{(i)}) 
%%   \triangleq \vg_{IS}(\vlambda)
%% \end{align*}
%% }%
%% %\vspace{-0.05in}
%% where \(\widetilde{w}\,(\vz) = p\,(\vz,\vx) / q_{\vlambda}(\vz)\) is known as the unnormalized importance weight, \(Z\) is the marginal \(p\left(\vx\right) = \int p\left(\vx, \vz\right) \, d\vz\), and \(\vz^{(1)},\, \ldots,\, \vz^{(N)}\) are \(N\) independent samples from \(q_{\vlambda}(\vz)\).
%% This scheme is equivalent to adaptive IS methods~\citep{bugallo_adaptive_2017} since the IS proposal \(q_{\vlambda}(\vz)\) is iteratively optimized based on the current samples.
%% Although IS is unbiased, it is numerically unstable because of \(Z\). %in \cref{eq:is_est}.
%% A more stable alternative is to use the normalized weight, which results in the self-normalized IS (SNIS) approximation.
%% Unfortunately, SNIS still fails to converge even on moderate dimensional objectives and unlike IS, it is no longer unbiased~\citep{robert_monte_2004}.

\vspace{-2ex}
\subsection{Markov Chain Gradient Descent}\label{section:mcgd}
\vspace{-1.0ex}
\paragraph{\textbf{Overview of MCGD}}
Markov chain gradient descent (MCGD, \citealt{duchi_ergodic_2012, NEURIPS2018_1371bcce}) is a family of algorithms that minimize a function \(f\) defined as \(f\left(\vlambda\right) = \int f\left(\vlambda, \veta\right) \, \Pi\left(d\veta\right)\), where \(\veta\) is random noise, and \(\Pi\left(d\veta\right)\) is its probability measure.
MCGD repeats the steps 
{%\small
\begin{align}
  \vlambda_{t+1}    = \vlambda_{t} - \gamma_t \, \vg\left(\vlambda_t, \veta_{t}\right),\quad 
  \rvveta_{t}  \sim P_{\vlambda_{t-1}}\left( \veta_{t-1}, \cdot \right),\label{eq:mcgd}
\end{align}
}%
where \(P_{\vlambda_{t-1}}\) is a \(\Pi\)-invariant Markov chain kernel that may depend on \(\vlambda_{t-1}\).
The noise of the gradient is Markovian and non-asymptotically biased, departing from vanilla SGD.
%Unlike vanilla SGD where the gradient noise is assumed to be \textit{i.i.d.} and unbiased, it is now .
Non-asymptotic convergence of this general algorithm has recently started to gather attention as by~\citet{duchi_ergodic_2012, NEURIPS2018_1371bcce, pmlr-v99-karimi19a, doan_finitetime_2020, doan_convergence_2020, debavelaere_convergence_2021}.

\vspace{-1.5ex}
\paragraph{\textbf{Applications of MCGD}}
MCGD encompasses an extensive range of problems, including distributed optimization~\citep{ram_incremental_2009}, reinforcement learning~\citep{tadic_asymptotic_2017, doan_convergence_2020, Xiong_Xu_Liang_Zhang_2021}, and expectation-minimization~\citep{pmlr-v99-karimi19a}, to name a few.
This paper extends this list with inclusive KL VI through the MCSA framework.

%%% Local Variables:
%%% TeX-master: "master"
%%% End:

\vspace{-1.5ex}
\section{Markov Chain Score Ascent}\label{section:mcsa}
\vspace{-1.0ex}

First, we develop Markov chain score ascent (MCSA), a framework for inclusive KL minimization with MCGD.
This framework will establish the connection between MSC/JSA and MCGD.

\vspace{-1.ex}
\subsection{Markov Chain Score Ascent as a Special Case of Markov Chain Gradient Descent}\label{section:convergence}
\vspace{-1.ex}
As shown in~\cref{eq:mcgd}, the basic ingredients of MCGD are the target function \(f\left(\vlambda, \eta\right)\), the gradient estimator \(\vg\left(\vlambda, \eta\right)\), and the Markov chain kernel \(P_{\vlambda}\left(\eta, \cdot\right)\).
Obtaining MCSA from MCGD boils down to designing \(\vg\) and \(P_{\vlambda}\) such that \(f\left(\vlambda\right) = \DKL{\pi}{q\left(\cdot; \vlambda\right)} \).
The following proposition provides sufficient conditions on \(\vg\) and \(P_{\vlambda}\) to achieve this goal.

\vspace{0.07in}

\begin{theoremEnd}{proposition}\label{thm:product_kernel}
  Let \(\veta = \left( \vz^{(1)}, \vz^{(2)}, \ldots, \vz^{(N)} \right)\) and a Markov chain kernel \(P_{\vlambda}\left(\veta, \cdot\right)\) be \(\Pi\)-invariant where \(\Pi\) is defined as
  {%\small
  \[
  \Pi\left(\veta\right) = \pi\left(\vz^{(1)}\right) \, \pi\left(\vz^{(2)}\right) \times \ldots \times \pi\left(\vz^{(N)}\right).
  \]
  }
  Then, by defining the objective function \(f\) and the gradient estimator \(\vg\) to be 
  {%\small
  \begin{align*}
    \textstyle
    f\left(\vlambda, \veta\right) =  -\frac{1}{N} \sum^{N}_{n=1} \log q\left(\vz^{(n)}; \vlambda\right) - \mathbb{H}\left[\,\pi\,\right] 
    \quad\text{and}\quad
    \vg\left(\vlambda, \veta\right) =  -\frac{1}{N} \sum^{N}_{n=1} \vs\left(\vz^{(n)}; \vlambda\right),
  \end{align*}
  }
  where \(\mathbb{H}\left[\,\pi\,\right]\) is the entropy of \(\pi\), MCGD results in inclusive KL minimization as
  {%\small
  \begin{align*}
    \Esub{\Pi}{ f\left(\vlambda, \rvveta\right) } = \DKL{\pi}{q\left(\cdot; \vlambda\right)}
    \quad\text{and}\quad
    \Esub{\Pi}{ \vg\left(\vlambda, \rvveta\right) } = \nabla_{\vlambda} \DKL{\pi}{q\left(\cdot; \vlambda\right)}.
  \end{align*}
  }
\end{theoremEnd}
\begin{proofEnd}
  For notational convenience, we define the shorthand
  \begin{align*}
    \pi\left(\vz^{(1:N)}\right) = \pi\left(\vz^{(1)}\right) \, \pi\left(\vz^{(2)}\right) \times \ldots \times \pi\left(\vz^{(N)}\right).
  \end{align*}
  Then,
  \begin{alignat}{2}
    &\Esub{\Pi}{ f\left(\vlambda, \rvveta\right) }
    \nonumber
    \\
    &\quad=
    \int \left(- \frac{1}{N} \sum^{N}_{n=1} \log q\left(\vz^{(n)}; \vlambda\right) - \mathbb{H}\left[\,\pi\,\right]\right) \, \pi\left(\vz^{(1:N)}\right) \, d\vz^{(1:N)}
    \nonumber
    \\
    &\quad=
    \int \left(- \frac{1}{N} \sum^{N}_{n=1} \log q\left(\vz^{(n)}; \vlambda\right)\right) \, \pi\left(\vz^{(1:N)}\right) \, d\vz^{(1:N)} - \mathbb{H}\left[\,\pi\,\right]
    \nonumber
    \\
    &\quad=
     \frac{1}{N} \sum^{N}_{n=1} \left\{ \int \big(-\log q\,(\,\vz^{(n)}; \vlambda\,)\,\big) \, \pi\left(\vz^{(1:N)}\right) \, d\vz^{(1:N)} \right\} - \mathbb{H}\,[\,\pi\,] 
    \nonumber
    \\
    &\quad=
    \frac{1}{N} \sum^{N}_{n=1} \int \big( -\log q\,(\,\vz^{(n)}; \vlambda\,) \,\big) \, \pi\left(\vz^{(n)}\right) \, d\vz^{(n)} - \mathbb{H}\,[\,\pi\,]
    &&\quad{\text{\textit{Marginalized \(\vz^{(m)}\) for all \(m \neq n\)}}}
    \nonumber
    \\
    &\quad=
    \frac{1}{N} \sum^{N}_{n=1} \int \big( -\log q\,(\,\vz^{(n)}; \vlambda\,) + \log\pi\left(\vz^{(n)}\right) \,\big) \, \pi\left(\vz^{(n)}\right) \, d\vz^{(n)}
    &&\quad{\text{\textit{Definition of \(\mathbb{H}\left[\pi\right]\)}}}
    \nonumber
    \\
    &\quad=
    \frac{1}{N} \sum^{N}_{n=1} \int \pi\left(\vz^{(n)}\right) \, \log \frac{\pi\left(\vz^{(n)}\right)}{q\,(\,\vz^{(n)}; \vlambda\,)}  \, d\vz^{(n)}
    \nonumber
    \\
    &\quad=
    \frac{1}{N} \sum^{N}_{n=1} \DKL{\pi}{q\left(\cdot; \vlambda\right)}
    \nonumber
    &&\quad{\text{\textit{Definition of \(d_{\text{KL}}\)}}}
    \\
    &\quad=
    \DKL{\pi}{q\left(\cdot; \vlambda\right)}.\label{eq:F_KL}
  \end{alignat}
  For \(\Esub{\Pi}{ \vg\left(\vlambda, \rvveta\right) }\), note that 
  \begin{align}
    \nabla_{\vlambda} f\left(\vlambda, \veta\right) = -\frac{1}{N} \sum^{N}_{n=1} \nabla_{\vlambda} \log q\left(\vz^{(n)}; \vlambda\right) = -\frac{1}{N} \sum^{N}_{n=1} \vs\left(\vz^{(n)}; \vlambda\right) = \vg\left(\vlambda, \veta\right).\label{eq:F_grad_G}
  \end{align}
  Therefore, it suffices to show that
  \begin{align*}
    \nabla_{\vlambda} \DKL{\pi}{q\left(\cdot; \vlambda\right)}
    &=
    \nabla_{\vlambda} \Esub{\Pi}{ f\left(\vlambda, \rvveta\right) }
    &&\text{\textit{\cref{eq:F_KL}}}
    \\
    &=
    \Esub{\Pi}{ \nabla_{\vlambda}  f\left(\vlambda, \rvveta\right) }
    &&\text{\textit{{Leibniz derivative rule}}}
    \\
    &=
    \Esub{\Pi}{ \vg\left(\vlambda, \rvveta\right) }.
    &&\text{\textit{\cref{eq:F_grad_G}}}
  \end{align*}
\end{proofEnd}

%%% Local Variables:
%%% TeX-master: "master"
%%% End:

\begin{table*}[t]
\vspace{-1ex}
\centering
\caption{Convergence Rates of MCGD Algorithms}\label{table:convergence}
\vspace{-0.05in}
\setlength{\tabcolsep}{3pt}
\begin{threeparttable}
  \begin{tabular}{lllcc}\toprule
    \multicolumn{1}{c}{\footnotesize\textbf{Algorithm}} & \multicolumn{1}{c}{\footnotesize\textbf{Stepsize Rule}} & \multicolumn{1}{c}{\footnotesize\textbf{Gradient Assumption}} & {\footnotesize\textbf{Rate}} & {\footnotesize\textbf{Reference}} \\\midrule
    \multirow{2}{*}{\small Mirror Descent\tnote{1}}
    & \multirow{2}{*}{\small\(\gamma_t = \gamma / \sqrt{t}\)}
    & \multirow{2}{*}{\small\(\E{ {\|\, \vg\left(\rvvlambda, \rvveta\right) \,\|}_*^2 \mid \mathcal{F}_{t-1} } < G^2\)}
    & \multirow{2}{*}{\small\(\mathcal{O}\left(\frac{G^2 \log T}{ \log \rho^{-1} \sqrt{T}}\right)\)}
    & {\footnotesize\citet{duchi_ergodic_2012}}
    \\
    &&&& {\footnotesize{Corollary 3.5}}
    \\\cdashlinelr{1-5}
    \multirow{2}{*}{\small SGD-Nesterov\tnote{2}}
    & {\small\(\gamma_t = 2/(t + 1)\)}
    & \multirow{2}{*}{\footnotesize\( {\|\,\vg\left(\rvvlambda, \rvveta\right)\|}_2 < G \)}
    & \multirow{2}{*}{\small\(\mathcal{O}\left(\frac{G^2 \log T}{ \sqrt{T}}\right)\)}
    & {\footnotesize\citet{doan_convergence_2020}}
    \\
    & {\footnotesize\(\beta_t = \frac{1}{2 \, L \sqrt{t + 1}}\)}
    &&& {\footnotesize{Theorem 2}}
    \\\cdashlinelr{1-5}
    \multirow{2}{*}{\small SGD\tnote{3}}
    & {\footnotesize\(\gamma_t = \gamma/t\)}
    & \multirow{2}{*}{\footnotesize\( {\|\,\vg\left(\rvvlambda, \rvveta\right)\|}_* < G \left( \norm{\vlambda}_2 + 1 \right) \)}
    & \multirow{2}{*}{\small\(\mathcal{O}\left(\frac{G^2 \log T}{ T}\right)\)}
    & {\footnotesize\citet{doan_finitetime_2020}}
    \\ 
    & {\footnotesize\(\gamma = \min\{\nicefrac{1}{2\,L}, \nicefrac{2 L}{\mu}\}\)}
    &&& {\footnotesize{Theorem 1,2}}
    \\ \bottomrule
  \end{tabular}
  \begin{tablenotes}[flushleft]
  \item[]{%
    \footnotesize\textbf{\textit{Notation}:}
    \(^1\)\(\mathcal{F}_t\) is the \(\sigma\)-field formed by the iterates \(\veta_t\) up to the \(t\)th MCGD iteration, \(\norm{\vx}_* \) is the dual norm of \(\vx\);
    \(^2\)\(\beta_t\) is the stepsize of the momentum;
    \(^2\)\(^3\)\(L\) is the Lipschitz smoothness constant;
    \(^3\)\(\mu\) is the strong convexity constant.
  }
  \end{tablenotes}
\end{threeparttable}
\vspace{-2ex}
\end{table*}

This simple connection between MCGD and VI paves the way toward the non-asymptotic analysis of JSA and MSC.
%We will later propose a third novel scheme that conforms to~\cref{thm:product_kernel}.
Note that \(N\) here can be regarded as the computational budget of each MCGD iteration since the cost of
\begin{enumerate*}[label=\textbf{(\roman*)}]
  \item generating the Markov chain samples \(\vz^{(1)}, \ldots, \vz^{(N)}\) and
  \item computing the gradient \(\vg\)
\end{enumerate*}
will linearly increase with \(N\).

In addition, the MCGD framework often assumes \(P\) to be geometrically ergodic.
An exception is the analysis of~\citet{debavelaere_convergence_2021} where they work with polynomially ergodic kernels.
\begin{assumption}{(Markov chain kernel)}\label{thm:kernel_conditions}
\vspace{-0.05in}
  The Markov chain kernel \(P\) is geometrically ergodic as
  {%\small
  \[
  \DTV{P_{\vlambda}^{n}\left(\veta, \cdot\right)}{ \Pi } \leq C \, \rho^{n}
  \]
  }
  for some positive constant \(C\).
\end{assumption}

\vspace{-2ex}
\subsection{Non-Asymptotic Convergence of Markov Chain Score Ascent}%\label{}
\vspace{-1.5ex}
\paragraph{Non-Asymptotic Convergence}
Through~\cref{thm:product_kernel},~\cref{thm:kernel_conditions} and some technical assumptions on the objective function, we can apply the existing convergence results of MCGD to MCSA.
\cref{table:convergence} provides a list of relevant results.
Apart from properties of the objective function (such as Lipschitz smoothness), the convergence rates are stated in terms of the gradient bound \(G\), kernel mixing rate \(\rho\), and the number of MCGD iterations \(T\).
We focus on \(G\) and \(\rho\) as they are closely related to the design choices of different MCSA algorithms.

\vspace{-0.1in}
\paragraph{Convergence and the Mixing Rate \(\rho\)}
\citet{duchi_ergodic_2012} was the first to provide an analysis of the general MCGD setting.
Their convergence rate is dependent on the mixing rate through the \(1 / \log \rho^{-1}\) term.
For MCSA, this result is overly conservative since, on challenging problems, mixing can be slow such that \(\rho \approx 1\).
Fortunately, \citet{doan_convergence_2020,doan_finitetime_2020} have recently shown that it is possible to obtain a rate independent of the mixing rate \(\rho\).
For example, in the result of~\citet{doan_finitetime_2020}, the influence of \(\rho\) decreases in a rate of \(\mathcal{O}\left(\nicefrac{1}{T^2}\right)\).
%The fact that the convergence rate can be independent of the mixing rate is critical.
This observation is critical since it implies that \textbf{trading a ``slower mixing rate'' for ``lower gradient variance'' could be profitable}.
We exploit this observation in our novel MCSA scheme in~\cref{section:pmcsa}.

\vspace{-0.1in}
\paragraph{Gradient Bound \(G\)}
Except for \citet{doan_finitetime_2020}, most results assume that the gradient is bounded for \(\forall\veta,\vlambda\) as {\footnotesize\( {\| \vg\left(\vlambda, \veta\right) \|} < G \)}.
Admittedly, this condition is strong, but it is similar to the bounded variance assumption {\footnotesize\(\mathbb{E}\,[\norm{\vg}^2]  < G^2\)} used in vanilla SGD, which is also known to be strong as it contradicts strong convexity~\citep{pmlr-v80-nguyen18c}.
Nonetheless, assuming \(G\) can have practical benefits beyond theoretical settings.
For example,~\citet{pmlr-v108-geffner20a} use \(G\) to compare the performance different VI gradient estimators.
In a similar spirit, we will obtain the gradient bound \(G\) of different MCSA algorithms and compare their theoretical performance.

\vspace{-0.1in}
\section{Demystifying Prior Markov Chain Score Ascent Methods}\label{section:comparison}
\vspace{-0.1in}
In this section, we will show that MSC and JSA both qualify as MCSA methods.
Furthermore, we establish 
\begin{enumerate*}[label=\textbf{(\roman*)}]
  \item the mixing rate of their implicitly defined kernel \(P\) and
  \item the upper bound on their gradient variance.
\end{enumerate*}
This will provide insight into their practical non-asymptotic performance.

\vspace{-1ex}
\subsection{Technical Assumptions}\label{section:assumption}
\vspace{-1ex}
To cast previous methods into MCSA, we need some technical assumptions.

\begin{assumption}{(Bounded importance weight)}\label{thm:bounded_weight}
  The importance weight ratio \(w\left(\vz\right) = \pi\left(\vz\right) / q\left(\vz; \vlambda\right)\) is bounded by some finite constant as \(w^* < \infty\) for all \(\vlambda \in \Lambda\) such that \(r = \left(1 - 1/w^*\right) < 1\).
\end{assumption}
\vspace{-0.05in}
This assumption is necessary to ensure~\cref{thm:kernel_conditions}, and can be practically ensured by using a variational family with heavy tails~\citep{NEURIPS2018_25db67c5} or using a defensive mixture~\citep{hesterberg_weighted_1995, holden_adaptive_2009} as
%\vspace{-0.07in}
{\[
  q_{\text{def.}}\left(\vz; \vlambda \right) = \alpha \, q\left(\vz; \vlambda\right) + (1 - \alpha) \, \nu\left(\vz\right),
\]}%
%\vspace{-0.01in}}%
where \(0 < \alpha < 1\) and \(\nu\left(\cdot\right)\) is a heavy tailed distribution such that \(\sup_{\vz \in \mathcal{Z}} \pi\left(\vz\right) / \nu\left(\vz\right) < \infty\).
Note that \(q_{\text{def.}}\) is only used in the Markov chain kernels and \(q\left(\cdot;\vlambda\right)\) is still the output of the VI procedure.
While these tricks help escape slowly mixing regions, this benefit quickly vanishes as \(\lambda\) converges.
Therefore, ensuring \cref{thm:bounded_weight} seems unnecessary in practice unless we absolutely care about ergodicity. (Think of the adaptive MCMC setting for example.~\citealt{holden_adaptive_2009, pmlr-v151-brofos22a}).

\vspace{-0.1in}
\paragraph{\textbf{Model (Variational Family) Misspecification and \(w^*\)}}
Note that \(w^*\) is bounded below exponentially by the inclusive KL as shown in~\cref{thm:wstar}.
Therefore, \(w^*\) will be large 
\begin{enumerate*}[label=\textbf{(\roman*)}]
    \item in the initial steps of VI and
    \item under model (variational family) misspecification.
\end{enumerate*}
%In practice, for practical it is fair to expect \(w^*\) will be large.

\begin{assumption}{(Bounded Score)}\label{thm:bounded_score}
  The score gradient is bounded for \(\forall \vlambda \in \Lambda\) and \(\forall \vz \in \mathcal{Z}\) such that \(\norm{\vs\left(\vlambda; \vz\right)}_2 \leq L \) for some finite constant \(L > 0\).
\end{assumption}
Although this assumption is strong, it enables us to compare the gradient variance of MCSA methods.
We empirically justify the bounds obtained using~\cref{thm:bounded_score} in \cref{section:simulation}.

%\pagebreak
\vspace{-0.07in}
\subsection{Markovian Score Climbing}
\vspace{-0.07in}

MSC (\cref{alg:msc} in \cref{section:pseudocode}) is a simple instance of MCSA where \(\veta_t = \vz_t\) and \(P_{\vlambda_t} = K_{\vlambda_t}\) is the conditional importance sampling (CIS) kernel (originally proposed by~\citet{andrieu_uniform_2018}) where the proposals are generated from \(q\left(\cdot; \vlambda_t\right)\).
Although MSC uses only a single sample for the Markov chain, the CIS kernel internally operates \(N-1\) proposals.
Therefore, \(N\) in MSC has a different meaning, but it still indicates the computational budget.

\begin{theoremEnd}[all end]{proposition}\label{thm:wstar}
The maximum importance weight
\(w^* = \sup_{\vz} w\left(\vz\right) = \sup_{\vz} \pi\left(\vz\right)/q\left(\vz;\vlambda\right)\)
is bounded below exponentially by the KL divergence as
\[
    \exp\left( \DKL{\pi}{q\left(\cdot; \vlambda\right)} \right) < w^*.
\]
\end{theoremEnd}
\begin{proofEnd}
  \begin{alignat*}{2}
    \DKL{\pi}{q\left(\cdot; \vlambda\right)}
    &=
    \Esub{\rvvz \sim \pi\left(\cdot\right)}{ \log \frac{\pi\left(\rvvz\right)}{q\left(\rvvz; \vlambda\right)} }
    &&\quad\text{\textit{Definition of \(d_{\text{KL}}\)}}
    \\
    &\leq
    \log \Esub{\rvvz \sim \pi\left(\cdot\right)}{ \frac{\pi\left(\rvvz\right)}{q\left(\rvvz; \vlambda\right)} }
    &&\quad\text{\textit{Jensen's inequality}}
    \\
    &\leq
    \log \Esub{\rvvz \sim \pi\left(\cdot\right)}{ w^* }
    \\
    &=
    \log w^*.
  \end{alignat*}
\end{proofEnd}

\begin{theoremEnd}[all end]{lemma}\label{thm:product_measure_bound}
  For the probability measures \(p_1, \ldots, p_N\) and \(q_1, \ldots, q_N\) defined on a measurable space \((\mathsf{X}, \mathcal{A})\) and an arbitrary set \(A \in \mathcal{A}\),
  \begin{align*}
    &\abs{
    \int_{A^N}
    p_1\left(dx_1\right)
    p_2\left(dx_2\right)
    \times
    \ldots
    \times
    p_N\left(dx_N\right)
    -
    q_1\left(dx_1\right)
    q_2\left(dx_2\right)
    \times
    \ldots
    \times
    q_N\left(dx_N\right)
  }
    \\
  &\qquad\leq
  \sum_{n=1}^N
  \abs{
    \int_{A}
    p_n\left(dx_n\right)
    -
    q_n\left(dx_n\right)
  }
  \end{align*}
\end{theoremEnd}
\begin{proofEnd}
  By using the following shorthand notations
  \begin{alignat*}{2}
    p_{(1:N)}\left(dx_{(1:N)}\right)
    &= 
    p_1\left(dx_1\right)
    p_2\left(dx_2\right)
    \times
    \ldots
    \times
    p_N\left(dx_N\right)
    \\
    q_{(1:N)}\left(dx_{(1:N)}\right)
    &= 
    q_1\left(dx_1\right)
    q_2\left(dx_2\right)
    \times
    \ldots
    \times
    q_N\left(dx_N\right),
  \end{alignat*}
  the result follows from induction as
  \begin{alignat}{2}
    &\abs{
      \int_{A^N}
      p_{(1:N)}\left(dx_{(1:N)}\right)
      -
      q_{(1:N)}\left(dx_{(1:N)}\right)
    }
    \nonumber
    \\
    &\quad=
    \Bigg|\;
    \left( \int_{A} p_1\left(dx_1\right) - q_1\left(dx_1\right) \right) \,
    \int_{A^{N-1}} p_{(2:N)}\left(dx_{(2:N)}\right)
    \nonumber
    \\
    &\qquad\quad+
    \int_{A} q_1\left(dx_1\right) \,
    {\left(
      \int_{A^{N-1}}
      p_{(2:N)}\left(dx_{(2:N)}\right)
      -
      q_{(2:N)}\left(dx_{(2:N)}\right)
    \right)}
    \;\Bigg|
    \nonumber
    \\
    &\quad\leq
    \Bigg|
    \int_{A} p_1\left(dx_1\right) - q_1\left(dx_1\right)
    \Bigg|\;
    \int_{A^{N-1}} p_{(2:N)}\left(dx_{(2:N)}\right)
    \nonumber
    \\
    &\qquad\quad+
    \int_{A} q_1\left(dx_1\right) \,
    {
    \Bigg|\;
      \int_{A^{N-1}}
      p_{(2:N)}\left(dx_{(2:N)}\right)
      -
      q_{(2:N)}\left(dx_{(2:N)}\right)
    }
    \;\Bigg|
    &&\quad\text{\textit{Triangle inequality}}
    \nonumber
    \\
    &\quad\leq
    \Bigg|
    \int_{A} p_1\left(dx_1\right) - q_1\left(dx_1\right)
    \Bigg|\;
    \nonumber
    \\
    &\qquad\quad+
    {
    \Bigg|\;
      \int_{A^{N-1}}
      p_{(2:N)}\left(dx_{(2:N)}\right)
      -
      q_{(2:N)}\left(dx_{(2:N)}\right)
    }
    \;\Bigg|.
    &&\quad\text{\textit{Applied \(p_n\left(A\right), q_n\left(A\right) \leq 1 \)}}
    \nonumber
\end{alignat}
\end{proofEnd}

\begin{theoremEnd}[all end]{lemma}\label{thm:second_moment_bound}
  Let \(\rvvg\) be a vector-valued, biased estimator of \(\vmu\), where the bias is denoted as \(\mathsf{Bias}\left[\rvvg\right] = \norm{ \mathbb{E}\rvvg - \vmu }_2\) and the mean-squared error is denoted as \(\mathsf{MSE}\left[\rvvg\right] = \mathbb{E}\norm{\rvvg - \vmu }_2^2\).
  Then, the second moment of \(\rvvg\) is bounded as
  \begin{enumerate}
  \item[\ding{182}]
  \(
    \mathbb{E}\norm{\rvvg}^2_2
    \leq
    \mathbb{V}{\rvvg} + {\mathsf{Bias}\left[\rvvg\right]}^2 + 2\, \mathsf{Bias}\left[\rvvg\right] \norm{\vmu}_2 + \norm{\vmu}^2_2,
  \)
  \item[\ding{183}]
  \(
    \mathbb{E}\norm{\rvvg}^2_2 
    \leq
    \mathsf{MSE}\left[{\rvvg}\right] + 2\, \mathsf{Bias}\left[\rvvg\right] \norm{\vmu}_2 + \norm{\vmu}^2_2,
  \)
  \end{enumerate}
  where \(\mathbb{V}\rvvg = \mathbb{E}\norm{ \rvvg - \mathbb{E}\vg}_2^2\) is the variance of the estimator.
\end{theoremEnd}
\begin{proofEnd}
  \ding{182} follows from the decomposition
  {%\small
  \begin{alignat}{2}
    \E{ \norm{\rvvg}^2_{2} }
    &\;=
    \mathbb{V}{ \rvvg }
    +
    \norm{ \mathbb{E}{ \rvvg } }_2^2
    \nonumber
    \\
    &\;=
    \mathbb{V}{ \rvvg }
    +
    \norm{
      \mathbb{E}{ \rvvg } - \vmu  + \vmu
    }_2^2
    \nonumber
    \\
    &\;=
    \mathbb{V}{ \rvvg }
    +
    \norm{
      \mathbb{E}{ \rvvg } - \vmu  
    }_2^2
    \nonumber
    +
    2 \, {\left(
      \mathbb{E}\rvvg - \vmu 
    \right)}^{\top}
    \vmu
    + 
    \norm{\vmu}^2_2
    &&\quad\text{\textit{Expanded quadratic}}
    \nonumber
    \\
    &\;\leq
    \mathbb{V}{ \rvvg }
    +
    \norm{
      \mathbb{E}{ \rvvg   } - \vmu 
    }_2^2
    +
    2 \,\norm{
      \mathbb{E}{ \rvvg  } - \vmu
    }_2
    \norm{\vmu}_2
    + 
    \norm{
      \vmu
    }_2^2
    \nonumber
    &&\quad\text{\textit{Cauchy-Shwarz inequality}}
    \\
    &\;=
    \mathbb{V}{ \rvvg  }
    +
    {\mathsf{Bias}\left[\rvvg\right]}^2
    +
    2 \,\mathsf{Bias}\left[\rvvg\right]
    \norm{\vmu}_2
    + 
    \norm{
      \vmu
    }_2^2.
    \nonumber
    &&\quad\text{\textit{Definition of bias}}
  \end{alignat}
  }
  Meanwhile, by the well-known bias-variance decomposition formula of the mean-squared error, \ding{183} directly follows from \ding{182}.
\end{proofEnd}

\vspace{0.04in}

\begin{theoremEnd}{theorem}\label{thm:msc}
  MSC~\citep{NEURIPS2020_b2070693} is obtained by defining 
  {%\small
  \begin{align*}
  P_{\lambda}^k\left(\veta, d\veta^{\prime}\right)
  = 
  K_{\lambda}^k\left(\vz, d\vz^{\prime}\right)
  \end{align*}
  }%
  with  \(\veta_t = \vz_t\), where \(K_{\vlambda}\left(\vz, \cdot\right)\) is the CIS kernel with \(q_{\text{def.}}\left(\cdot; \vlambda\right)\) as its proposal distribution.
  Then, given~\cref{thm:bounded_weight,thm:bounded_score}, the mixing rate and the gradient bounds are given as
  {%\small
  \begin{align*}
    \textstyle
  \DTV{P_{\vlambda}^k\left(\veta, \cdot\right)}{\Pi} \leq  {\left(1 - \frac{N - 1}{2 w^* + N - 2}\right)}^k\quad \text{and}\quad
  {\small
  \E{ {\lVert\rvvg_{t,\text{MSC}}\rVert}^2 \,\middle|\, \mathcal{F}_{t-1} } \leq  L^2,
  }
  \end{align*}
  }%
  where \(w^* = \sup_{\vz} \pi\left(\vz\right) / q_{\text{def.}}\left(\vz;\vlambda\right)\).
\end{theoremEnd}
\begin{proofEnd}
  MSC is described in~\cref{alg:msc}. 
  At each iteration, it performs a single MCMC transition with the CIS kernel where it internally uses \(N\) proposals.
  That is,
  \begin{align*}
    \rvvz_t \mid \rvvz_{t-1}, \rvvlambda_{t-1} &\sim K_{\rvvlambda_{t-1}}\left(\rvvz_{t-1}, \cdot\right)
    \\
    \rvvg_{t,\text{MSC}} &= -\vs\left(\vlambda, \rvvz_t\right),
  \end{align*}
  where \(K_{\rvvlambda_{t-1}}\) is the CIS kernel using \(q_{\text{def.}}\left(\cdot; \rvvlambda_{t-1}\right)\).

  \paragraph{Ergodicity of the Markov Chain}
  The ergodic convergence rate of \(P_{\vlambda}\) is equal to that of \(K_{\vlambda}\), the CIS kernel proposed by~\citet{NEURIPS2020_b2070693}. 
  Although not mentioned by~\citet{NEURIPS2020_b2070693}, this kernel has been previously proposed as the iterated sequential importance resampling (i-SIR) by \citet{andrieu_uniform_2018} with its corresponding geometric convergence rate as
  \begin{alignat*}{2}
    \DTV{P^{k}_{\vlambda}\left(\veta, \cdot\right)}{\Pi}
    =
    \DTV{K^{k}_{\vlambda}\left(\vz, \cdot\right)}{\pi}
    \leq
    {\left(1 - \frac{N - 1}{2 w^* + N - 2}\right)}^k.
  \end{alignat*}

  \paragraph{\textbf{Bound on the Gradient Variance}}
  The bound on the gradient variance is straightforward given \cref{thm:bounded_score}.
  For simplicity, we denote the rejection state as \(\vz^{(1)}_t = \vz_{t-1} \).
  Then,
  \begin{alignat}{2}
    &\E{ {\lVert\rvvg_{t,\text{MSC}}\rVert}^2 \,\middle|\, \mathcal{F}_{t-1} }
    \nonumber
    \\
    &\;=
    \Esub{\rvvz_t \sim K_{\rvvlambda_{t-1}}\left(\rvvz_{t-1}, \cdot\right)}{
      \norm{ \vs\left(\vlambda; \rvvz_t\right) }^2 \,\middle|\,
      \rvvlambda_{t-1}, \rvvz_{t-1}
    }
    \nonumber
    \\
    &\;=
    \int
    \sum^{N}_{n=1}
    \frac{
      w\left(\vz^{(n)}_t\right)
    }{
      \sum^{N}_{m=1} w\left(\vz^{(m)}_t\right)
    }
    \norm{ \vs\left(\vlambda; \vz^{(n)}_t\right) }^2
    \prod^{N}_{n=2}
    q\left(d\vz^{(n)}_t; \rvvlambda_{t-1}\right)
    \nonumber
    &&\quad\text{\textit{\citet{andrieu_uniform_2018}}}
    \\
    &\;\leq
    L^2 \,
    \int
    \sum^{N}_{n=1}
    \frac{
      w\left(\vz^{(n)}_t\right)
    }{
      \sum^{N}_{m=1} w\left(\vz^{(m)}_t\right)
    }
    \prod^{N}_{n=2}
    q\left(d\vz^{(n)}_t; \rvvlambda_{t-1}\right)
    \nonumber
    &&\quad\text{\textit{\cref{thm:bounded_score}}}
    \\
    &\;=
    L^2 \,
    \int
    \prod^{N}_{n=2}
    q\left(d\vz^{(n)}; \rvvlambda_{t-1}\right)
    \nonumber
    \\
    &\;=
    L^2.
    \nonumber
  \end{alignat}

\end{proofEnd}

%%% Local Variables:
%%% TeX-master: "master"
%%% End:

\vspace{-0.1in}
\paragraph{Discussion}
\cref{thm:msc} shows that the gradient variance of MSC is insensitive to \(N\).
Although the mixing rate does improve with \(N\), when \(w^*\) is large due to model misspecification and lack of convergence (see the discussion in \cref{section:assumption}), this will be marginal.
Overall, \textit{the performance of MSC cannot be improved by increasing the computational budget \(N\)}.

\vspace{-0.1in}
\paragraph{Rao-Blackwellization}
Meanwhile, \citeauthor{NEURIPS2020_b2070693} also provide a Rao-Blackwellized version of MSC we denote as MSC-RB.
Instead of selecting a single \(\vz_t\) by resampling over the \(N\) internal proposals, they suggest forming an importance-weighted estimate~\citep{robert_monte_2004}.
The theoretical properties of this estimator have been concurrently analyzed by~\citet{cardoso_brsnis_2022}.

\begin{theoremEnd}[all end]{lemma}\label{thm:iw_variance}
  Let the importance weight be defined as \(w\left(\vz\right) = \pi\left(\vz\right)/q\left(\vz\right)\).
  The variance of the importance weights is related to the \(\chi^2\) divergence as
  \begin{align*}
    \mathbb{V}_{q}{w\left(\rvvz\right)}
    =
    \DChi{\pi}{q}.
  \end{align*}
\end{theoremEnd}
\begin{proofEnd}
  {
  \begin{alignat}{2}
    \mathbb{V}_{q}{w\left(\rvvz\right)}
    =
    \Esub{q}{
      {\left(
        w\left(\rvvz\right)
      -
      \Esub{q}{
        w\left(\rvvz\right)
      }
      \right)}^2
    }
    =
    \Esub{q}{
      {\left(
      w\left(\rvvz\right)
      -
      1
      \right)}^2
    }
    =
    \int {\left( \frac{\pi\left(\vz\right)}{q\left(\vz\right)}  - 1 \right)}^2 q\left(d\vz\right)
    =
    \DChi{\pi}{q}.
    \nonumber
  \end{alignat}
  }
\end{proofEnd}

\begin{theoremEnd}{theorem}{\citep{cardoso_brsnis_2022}}\label{thm:mscrb}
  The gradient variance of MSC-RB is bounded as
  {%\small
  \begin{align*}
  {\textstyle
    \E{ {\lVert \rvvg_{t,\mathrm{MSC\text{-}RB}} \rVert}_2^2  \,\middle|\, \mathcal{F}_{t-1} }
    \leq
    4 \, L^2 \,\left[
    \frac{1}{N-1}\,\DChi{\pi}{q\left(\cdot; \rvvlambda_{t-1}\right)} 
    +
    \mathcal{O}\left(N^{-3/2} + \nicefrac{\gamma^{t-1}}{N-1}\right)
    \right]
    +
    {\lVert\vmu\rVert}^2_2,
  }
  \end{align*}
  }%
  where \(\vmu = \mathbb{E}_{\pi}\vs\left(\vlambda;\rvvz\right)\), \(\gamma = 2 w^* / \left(2 w^* + N - 2\right)\) is the mixing rate of the Rao-Blackwellized CIS kernel, and {\small\(
  \DChi{\pi}{q} = \int {\left(\nicefrac{\pi}{q} - 1\right)}^2 q\left(d\vz\right)
  \)} is the \(\chi^2\) divergence.
\end{theoremEnd}
\begin{proofEnd}
  Rao-Blackwellization of the CIS kernel is to reuse the importance weights \(w\left(\vz\right) = \pi\left(\vz\right) / q_{\text{def.}}\left(\vz\right)\) internally used by the kernel when forming the estimator.
  That is, the gradient is estimated as
  \begin{align*}
    &\rvvz^{(n)}_t \mid \rvvlambda_{t-1} \sim q\left(\cdot; \rvvlambda_{t-1}\right) \\
    &\rvvg_{t,\text{MSC-RB}} = -\sum^{N}_{n=2} \frac{w\left(\rvvz^{(n)}_t\right)}{\sum^{N}_{m=2}w\left(\rvvz^{(n)}_t\right) + w\left(\rvvz_{t-1}\right)} \vs\left(\rvvz^{(n)}_t\right) + \frac{w\left(\rvvz_{t-1}\right)}{\sum^{N}_{m=2}w\left(\rvvz^{(n)}_t\right) + w\left(\rvvz_{t-1}\right)} \vs\left(\rvvz_{t-1}\right).
  \end{align*}

  By \cref{thm:second_moment_bound}, the second moment of the gradient is bounded as
  \begin{alignat}{2}
    \E{ {\parallel \rvvg_{t,\text{MSC-RB}} \parallel}_2^2 \,\middle|\, \mathcal{F}_{t-1} } 
    &=
    \mathsf{MSE}\left[ \rvvg_{t, \text{MSC-RB}} \,\middle|\, \mathcal{F}_{t-1} \right]
    +
    2 \, {\mathsf{Bias}\left[ \rvvg_{t, \text{MSC-RB}} \,\middle|\, \mathcal{F}_{t-1} \right]}^{\top} \norm{\vmu}_2
    + \norm{\vmu}^2_2
    \nonumber
    \\
    &\leq
    \mathsf{MSE}\left[ \rvvg_{t, \text{MSC-RB}} \,\middle|\, \mathcal{F}_{t-1} \right]
    +
    2 \, L \, {\mathsf{Bias}\left[ \rvvg_{t, \text{MSC-RB}} \,\middle|\, \mathcal{F}_{t-1} \right]}
    + \norm{\vmu}^2_2.
    \label{eq:mscrb_decomp}
  \end{alignat}
  
  \citet[Theorem 3]{cardoso_brsnis_2022} show that the mean-squared error of this estimator, which they call bias reduced self-normalized importance sampling, is bounded as
  \begin{align*}
    \mathsf{MSE}\left[ \rvvg_{t,\text{MSC-RB}} \,\middle|\, \mathcal{F}_{t-1} \right] \leq
    4\,L^2 \, \Big[\,&
     \left( 1 + \epsilon^2 \right)\, \frac{1}{N-1}\,\Vsub{\rvvz \sim q_{\text{def.}}\left(\cdot; \rvvlambda_{t-1}\right)}{w\left(\rvvz\right) \mid \rvvlambda_{t-1}} 
     \\
     &+
     \left(1 + \epsilon^{-2} \right)\, \frac{1}{N^2} {\left(1 + w^*\right)}^2 
    \,\Big],
  \end{align*}
  for some arbitrary constant \(\epsilon^2\).
  The first term is identical to the variance of an \(N-1\)-sample self-normalized importance sampling estimator~\citep{10.1214/17-STS611}, while the second term is the added variance due to ``rejections.''

  Since the variance of the importance weights is well known to be related to the \(\chi^2\) divergence,
  \begin{alignat}{2}
    \mathsf{MSE}\left[ \rvvg_{t,\text{MSC-RB}} \,\middle|\, \mathcal{F}_{t-1} \right] \leq
    4\,L^2 \, \Big[\,&
     \left( 1 + \epsilon^2 \right)\, \frac{1}{N-1}\,\DChi{\pi}{q\left(\cdot; \rvvlambda_{t-1}\right)}
    \nonumber
     \\
     &+
     \left(1 + \epsilon^{-2} \right)\, \frac{1}{N^2} {\left(1 + w^*\right)}^2 
    \,\Big].
    &&\quad\text{\textit{\cref{thm:iw_variance}}}
    \nonumber
  \end{alignat}

  For \(\epsilon^2\), \citeauthor{cardoso_brsnis_2022} choose \(\epsilon^2 = {\left(N-1\right)}^{-1/2}\), which results in their stated bound
  \begin{align*}
    \mathsf{MSE}\left[ \rvvg_{t,\text{MSC-RB}} \,\middle|\, \mathcal{F}_{t-1} \right] \leq
    4\,L^2 \, \Big[\,
      \frac{1}{N-1}\,\DChi{\pi}{q\left(\cdot; \rvvlambda_{t-1}\right)}
      +
      \mathcal{O}\left(N^{-3/2}\right)
    \,\Big].
  \end{align*}

  Furthermore, they show that the bias term is bounded as
  \begin{align*}
    \mathsf{Bias}\left[ \rvvg_{t,\text{MSC-RB}} \right]
    \leq
    \frac{4 \, L}{N-1} \left( \DChi{\pi}{q\left(\cdot; \rvvlambda_{t-1}\right)} + 1 + w^*\right) {\left( \frac{2 w^*}{ 2 w^* + N - 2 } \right)}^{t-1}.
  \end{align*}
  Combining both the bias and the mean-squared error to \cref{eq:mscrb_decomp}, we obtain the bound
  \begin{align*}
    &\E{ {\lVert \rvvg_{t,\text{MSC-RB}} \rVert}_2^2 \,\middle|\, \mathcal{F}_{t-1} }
    \\
    &\;\leq 
    4\,L^2\,\bigg[
    \frac{1}{N-1}\,\DChi{\pi}{q\left(\cdot; \rvvlambda_{t-1}\right)}
    \\
    &\qquad+
    \frac{1}{N-1} \left( \DChi{\pi}{q\left(\cdot; \rvvlambda_{t-1}\right)} + 1 + w^*\right) {\left( \frac{2 w^*}{ 2 w^* + N - 2 } \right)}^{t-1}
    +
    \mathcal{O}\left(N^{-3/2}\right)
    \bigg]
    \\
    &=
    4 \, L^2 \,\bigg[
    \frac{1 + \gamma^{t-1}}{N-1}\,\DChi{\pi}{q\left(\cdot; \rvvlambda_{t-1}\right)} + \frac{\gamma^{t-1}}{N-1} + \frac{\gamma^{t-1} w^*}{N-1}
    +
    \mathcal{O}\left(N^{-3/2}\right)
    \bigg].
  \end{align*}
\end{proofEnd}

%%% Local Variables:
%%% TeX-master: "master"
%%% End:

The variance of MSC-RB decreases as \(\mathcal{O}\left(\nicefrac{1}{N-1}\right)\), which is more encouraging than vanilla MSC.
However, the first term depends on the \(\chi^2\) divergence, which is bounded below exponentially by the KL divergence~\citep{10.1214/17-STS611}.
Therefore, \textit{the variance of MSC-RB will be large on challenging problems where the \(\chi^2\) divergence is large}, although linear variance reduction is possible.

\vspace{-0.1in}
\subsection{Joint Stochastic Approximation}
\vspace{-0.1in}
JSA (\cref{alg:jsa} in \cref{section:pseudocode}) was proposed for deep generative models where the likelihood factorizes into each datapoint.
Then, subsampling can be used through a random-scan version of the independent Metropolis-Hastings (IMH, \citealt{hastings_monte_1970}) kernel.
Instead, we consider the general version of JSA with a vanilla IMH kernel since it can be used for any type of likelihood.
At each MCGD step, JSA performs multiple Markov chain transitions and estimates the gradient by averaging all the intermediate states, which is closer to how traditional MCMC is used.

\vspace{-0.1in}
\paragraph{Independent Metropolis-Hastings}
%A key element of JSA~\citep{pmlr-v124-ou20a} is that it uses the IMH kernel.
Similarly to MSC, the IMH kernel in JSA generates proposals from \(q\left(\cdot; \vlambda_t\right)\).
To show the geometric ergodicity of the implicit kernel \(P\), we utilize the geometric convergence rate of IMH kernels provided by~\citet[Theorem 2.1]{10.2307/2242610} and~\citet{wang_exact_2022}.
The gradient variance, on the other hand, is difficult to analyze, especially the covariance between the samples.
However, we show that, even if we ignore the covariance terms, the variance reduction with respect to \(N\) is severly limited in the large \(w^*\) regime.
To do this, we use the exact \(n\)-step marginal IMH kernel derived by~\citet{Smith96exacttransition} as
{
  \begin{align}
  K^n_{\vlambda}\left(\vz, d\vz^{\prime}\right) 
  = T_n\left(\, w\left(\vz\right) \vee w\left(\vz^{\prime}\right)\,\right) \, \pi\left(\vz^{\prime}\right) \, d\vz^{\prime}
  + \lambda^n\left(w\left(\vz\right)\right) \, \delta_{\vz}\left(d\vz^{\prime}\right),
  \label{eq:imh_exact_kernel}
  \end{align}
}%
where {\(w\left(\vz\right) = \pi\left(\vz\right)/q_{\text{def.}}\left(\vz; \vlambda\right)\), \(x \vee y = \max\left(x, y\right)\)}, and for {\(R\left(v\right) = \{\, \vz^{\prime} \mid w\,\left(\vz^{\prime}\right) \leq v \,\}\)}, 
{%\small
  \begin{align}
    %\textstyle
    T_n\left(w\right)      = \int_w^{\infty}
    \frac{n}{v^2}
    %\left(n / v^2\right)
    \, \lambda^{n-1}\left(v\right)\,dv
    \quad\text{and}\quad
    \lambda\left(w\right) =
    \int_{R\left(w\right)}
    \left( 1 - \frac{w\left(\vz^{\prime}\right)}{w}  \right)
    %\left( 1 - w\left(\vz^{\prime}\right)/w  \right)
    \pi\left(d\vz^{\prime}\right).\label{eq:T_lambda}
  \end{align}
}%

\begin{theoremEnd}[all end]{lemma}\label{thm:lambda_bound}
  For \(w^* = \sup_{\vz} w\left(\vz\right) \), \(\lambda\left(\cdot\right)\) in~\cref{eq:T_lambda} is bounded as
  \[
   \max\left(1 - \frac{1}{w}, 0\right) \leq \lambda\left(w\right) \leq 1 - \frac{1}{w^*}.
  \]
\end{theoremEnd}
\begin{proofEnd}
  The proof can be found in the proof of Theorem 3 of \citet{Smith96exacttransition}.
\end{proofEnd}

\begin{theoremEnd}[all end]{lemma}\label{thm:tn_bound}
  For \(w^* = \sup_{\vz} w\left(\vz\right) \), \(T_n\left(\cdot\right)\) in~\cref{eq:T_lambda} is bounded as
  \[
  T_n\left( w \right) \leq \frac{n}{w} \, {\left(1 - \frac{1}{w^*}\right)}^{n-1}.
  \]
\end{theoremEnd}
\begin{proofEnd}
  \begin{alignat*}{2}
    T_n\left(w\right) 
    &= \int_w^{\infty} \frac{n}{v^2} \, \lambda^{n-1}\left(v\right)\,dv
    &&\quad\text{\textit{\cref{eq:T_lambda}}}
    \\
    &\leq \int_w^{\infty} \frac{n}{v^2} \, {\left(1 - \frac{1}{w^*}\right)}^{n-1}\,dv
    &&\quad\text{\textit{\cref{thm:lambda_bound}}}
    \\
    &= n \, {\left(1 - \frac{1}{w^*}\right)}^{n-1}  \int_w^{\infty} \frac{1}{v^2} \,dv
    \\
    &= n \, {\left(1 - \frac{1}{w^*}\right)}^{n-1}  \left( {-\left.\frac{1}{v}\right\rvert^{\infty}_{w}} \right)
    \\
    &= \frac{n}{w} \, {\left(1 - \frac{1}{w^*}\right)}^{n-1}.
  \end{alignat*}
\end{proofEnd}

\begin{theoremEnd}[all end]{lemma}\label{thm:imh_expecation}
  For a positive test function \(f : \mathcal{Z} \rightarrow \mathbb{R}^{+}\), the estimate of a \(\pi\)-invariant independent Metropolis-Hastings kernel \(K\) with a proposal \(q\) is bounded as
  \begin{align*}
    \Esub{K^n\left(\vz, \cdot\right)}{ f \,\middle|\, \rvvz }
    \leq
    n \, r^{n-1} 
    \Esub{q}{f}
    +
    {r}^n \, f\left(\rvvz\right),
    %% \leq
    %% n \, \left(
    %% \Esub{q}{f}
    %% +
    %% \frac{1}{n} \, f\left(\vz\right)
    %% \right) 
  \end{align*}
  where \(w\left(\vz\right) = \pi\left(\vz\right) / q\left(\vz\right)\) and \(r = 1 - \nicefrac{1}{w^*}\) for \(w^* = \sup_{\vz} w\left(\vz\right) \).
\end{theoremEnd}
\begin{proofEnd}
  \begin{alignat*}{2}
    &\Esub{K^n\left(\vz, \cdot\right)}{ f \,\middle|\, \vz }
    \\
    &\quad=
    \int T_n\left(w\left(\vz\right) \vee w\left(\vz^{\prime}\right)\right) \, f\left(\vz^{\prime}\right) \, \pi\left(\vz^{\prime}\right) d\vz^{\prime}
    +
    \lambda^{n}\left(w\left(\rvvz\right)\right) \, f\left(\rvvz\right)
    &&\quad{\text{\textit{\cref{eq:imh_exact_kernel}}}}
    \\
    &\quad\leq
    \int \frac{n}{w\left(\vz\right) \vee w\left(\vz^{\prime}\right)} \, {\left(1 - \frac{1}{w^*}\right)}^{n-1} \, f\left(\vz^{\prime}\right) \, \pi\left(\vz^{\prime}\right) d\vz^{\prime}
    +
    \lambda^{n}\left(w\left(\rvvz\right)\right) \, f\left(\rvvz\right)
    &&\quad{\text{\textit{\cref{thm:tn_bound}}}}
    \\
    &\quad\leq
    \int \frac{n}{w\left(\vz^{\prime}\right)} \, {\left(1 - \frac{1}{w^*}\right)}^{n-1} \, f\left(\vz^{\prime}\right) \, \pi\left(\vz^{\prime}\right) d\vz^{\prime}
    +
    \lambda^{n}\left(w\left(\rvvz\right)\right) \, f\left(\rvvz\right)
    &&\quad{\frac{1}{w\left(\vz\right) \vee w\left(\vz^{\prime}\right)} \leq \frac{1}{w\left(\vz^{\prime}\right)}}
    \\
    &\quad=
    n \, {\left(1 - \frac{1}{w^*}\right)}^{n-1} \, 
    \int \frac{1}{w\left(\vz^{\prime}\right)} \, f\left(\vz^{\prime}\right) \, \pi\left(\vz^{\prime}\right) d\vz^{\prime}
    +
    \lambda^{n}\left(w\left(\rvvz\right)\right) \, f\left(\rvvz\right)
    \\
    &\quad=
    n \, {\left(1 - \frac{1}{w^*}\right)}^{n-1} \, 
    \int f\left(\vz^{\prime}\right) \, q\left(\vz^{\prime}\right) d\vz^{\prime}
    +
    \lambda^{n}\left(w\left(\rvvz\right)\right) \, f\left(\rvvz\right)
    &&\quad{\text{\textit{Definition of \(w\left(\vz\right)\)}}}
    \\
    &\quad\leq
    n \, {\left(1 - \frac{1}{w^*}\right)}^{n-1} \, 
    \int f\left(\vz^{\prime}\right) \, q\left(\vz^{\prime}\right) d\vz^{\prime}
    +
    {\left(1 - \frac{1}{w^*}\right)}^{n} \, f\left(\rvvz\right)
    &&\quad{\text{\textit{\cref{thm:lambda_bound}}}}
    \\
    &\quad=
    n \, {\left(1 - \frac{1}{w^*}\right)}^{n-1} 
    \Esub{q}{f}
    +
    {\left(1 - \frac{1}{w^*}\right)}^n \, f\left(\rvvz\right).
  \end{alignat*}
\end{proofEnd}

\begin{theoremEnd}[all end]{lemma}\label{thm:mcmc_bias}
  Let a \(\Pi\)-invariant Markov chain kernel \(P\) be geometrically ergodic as
  \begin{align*}
    \DTV{P^{n}\left(\veta_0, \cdot\right)}{ \Pi } \leq C \, \rho^{n}.
  \end{align*}
  Furthermore, let \(\widehat{\rvvg} = \vg\left(\rvveta\right)\) with \(\rvveta \sim P^n\left(\veta_0, \cdot\right)\) be the estimator of \(\mathbb{E}_{\Pi}\rvvg\) for some function \(\vg : \mathrm{H} \rightarrow \mathbb{R}^{D}\) bounded as \( \norm{\rvvg}_{2} \leq L \).
  The bias of \(\rvvg\), defined as
  \(
    \mathsf{Bias}\left[ \widehat{\rvvg} \right] = \mathbb{E}\parallel \widehat{\rvvg} - \mathbb{E}_{\pi}\rvvg \parallel
  \),
  is bounded as
  \begin{alignat*}{2}
    \mathsf{Bias}\left[ \hat{\rvvg} \right]
    \leq
    2 \, \sqrt{D} \, L \, C \, \rho^{n}.
  \end{alignat*}
\end{theoremEnd}
\begin{proofEnd}
  \begin{alignat}{2}
    \mathsf{Bias}\left[ \hat{\rvvg} \right]
    &=
    \norm{
      \mathbb{E}_{P^n\left(\veta_0, \cdot\right)}\vg
      -
      \mathbb{E}_{\Pi}\vg
    }_2
    \nonumber
    \\
    &\leq
    \sqrt{D}\,\norm{
      \mathbb{E}_{P^n\left(\veta_0, \cdot\right)}\vg
      -
      \mathbb{E}_{\Pi}\vg
    }_{\infty}
    &&\quad\text{\textit{for  \(\vx \in \mathbb{R}^D\), \(\norm{\vx}_2 \leq \sqrt{D}\norm{\vx}_{\infty}\)}}
    \nonumber
    \\
    &\leq
    \sqrt{D}\,L\,\sup_{\abs{h} \leq 1}
    \abs{
      \mathbb{E}_{P^n\left(\veta_0, \cdot\right)}h
      -
      \mathbb{E}_{\Pi}h
    }
    &&\quad\text{\textit{\(\norm{\vg}_{\infty} \leq \norm{\vg}_2 \leq L\)}}
    \nonumber
    \\
    &=
    2 \,  \sqrt{D} \, L \,
    \DTV{P^{n}\left(\veta^{(0)}, \cdot\right)}{\pi}
    &&\quad\text{\textit{Definition of \(d_{\mathrm{TV}}\)}}
    \nonumber
    \\
    &=
    2 \, \sqrt{D} \, L \, C \, \rho^{n}.
    &&\quad\text{\textit{Geometric ergodicity}}
    \nonumber
  \end{alignat}
\end{proofEnd}

%%% Local Variables:
%%% TeX-master: "master"
%%% End:

\begin{theoremEnd}{theorem}\label{thm:jsa}
  JSA~\citep{pmlr-v124-ou20a} is obtained by defining 
  {%\small
  \begin{align*}
  P_{\vlambda}^k\left(\veta, d\veta^{\prime}\right)
  = 
  K_{\vlambda}^{N\,\left(k-1\right) + 1}\left(\vz^{(1)}, d\vz^{\prime\;(1)}\right)
  \,
  K_{\vlambda}^{N\,\left(k-1\right) + 2}\left(\vz^{(2)}, d\vz^{\prime\;(2)}\right)
  \cdot
  \ldots 
  \cdot
  K_{\vlambda}^{N\,\left(k-1\right) + N}\left(\vz^{(N)}, d\vz^{\prime\;(N)}\right)
  \end{align*}
  }%
  with \(\veta_t = \big(\vz_t^{(1)}, \vz_t^{(2)}, \ldots, \vz_t^{(N)}\big)\).
  Then, given~\cref{thm:bounded_weight,thm:bounded_score}, the mixing rate and the gradient variance bounds are
  {\small
  \begin{align*}
   % \textstyle
    \DTV{P_{\vlambda}^k\left(\veta, \cdot\right)}{\Pi}
    \leq
    C\left(r, N\right)\,{r}^{k\,N}
    \;\;\text{and}\;\;
    \E{ {\lVert\vg_{t,\mathrm{JSA}}\rVert}^2_2 \,\middle|\, \mathcal{F}_{t-1} }
    \leq
    L^2 \,
    \left[\,
    \frac{1}{2} + \frac{3}{2}\,\frac{1}{N}
    + \mathcal{O}\left(\nicefrac{1}{w^* + r^{t\,N}}\right)
    \,\right]
    +
    C_{\text{cov}}
    +
    {\lVert\vmu\rVert}^2_2,
    \nonumber
  \end{align*}
  }%
  where
  \(\vmu = \mathbb{E}_{\pi}\vs\left(\vlambda;\rvvz\right)\), 
  \(\small\textstyle
  C_{\text{cov}} = \frac{2}{N^2} \sum^{N}_{n=2} \sum^{n-1}_{m = 1} \Cov{\vs\left(\vlambda; \rvvz^{(n)}_t\right), \vs\left(\vlambda; \rvvz^{(m)}_t\right) \,\middle|\, \mathcal{F}_{t-1} }
  \) is the sum of the covariance between the samples, \(w^* = \sup_{\vz} \pi\left(\vz\right) / q_{\text{def.}}\left(\vz;\vlambda\right)\), and \(C\left(r, N\right) > 0\) is a finite constant.
\end{theoremEnd}
\begin{proofEnd}

  JSA is described in~\cref{alg:jsa}. 
  At each iteration, it performs \(N\) MCMC transitions and uses the \(N\) intermediate states to estimate the gradient.
  That is,
  \begin{align*}
    \rvvz^{(1)}_{t} &\mid \rvvz_{t-1}^{(N)},\, \rvvlambda_{t-1} \sim K_{\rvvlambda_{t-1}}\left(\rvvz_{t-1}^{(N)}, \cdot\right) \\
    \rvvz^{(2)}_{t} &\mid \rvvz^{(1)}_{t},\, \rvvlambda_{t-1}  \sim K_{\rvvlambda_{t-1}}\left(\rvvz_{t}^{(1)}, \cdot\right) \\
    &\qquad\qquad\vdots
    \\
    \rvvz^{(N)}_{t} &\mid \rvvz^{(N-1)}_{t},\, \rvvlambda_{t-1}  \sim K_{\rvvlambda_{t-1}}\left(\rvvz_{t}^{(N-1)}, \cdot\right)
    \\
    \rvvg_{t,\text{JSA}}  &= -\frac{1}{N}\,\sum^{N}_{n=1} \, \vs\left(\vlambda, \rvvz^{(n)}_t\right),
  \end{align*}
  where \(K_{\rvvlambda_{t-1}}^n\) is an \(n\)-transition IMH kernel using \(q_{\text{def.}}\left(\cdot; \rvvlambda_{t-1}\right)\).
  Under \cref{thm:bounded_weight}, an IMH kernel is uniformly geometrically ergodic~\citep{10.2307/2242610, wang_exact_2022} as
  \begin{align}
    \DTV{K_{\vlambda}^k\left(\vz, \cdot\right)}{\pi} \leq r^{k}\label{eq:imh_ergodicity}
  \end{align}
  for any \(\vz \in \mathcal{Z}\).

  \paragraph{Ergodicity of the Markov Chain}
  The state transitions of the Markov chain samples \(\vz^{(1:N)}\) are visualized as 
  {\small
  \begin{center}
  \bgroup
  \setlength{\tabcolsep}{3pt}
  \def\arraystretch{1.8}
  \begin{tabular}{c|ccccc}
   & \(\vz^{(1)}_t\) & \(\vz^{(2)}_t\) & \(\vz^{(3)}_t\) & \(\ldots\) &  \(\vz^{(N)}_t\) \\ \midrule
   \(t=1\) & \(K_{\vlambda_1}\left(\vz_0, d\vz_1^{(1)}\right)\) & \(K_{\vlambda_1}^2\left(\vz_0, d\vz_1^{(2)}\right)\) & \(K_{\vlambda_1}^3\left(\vz_0, d\vz_1^{(3)}\right)\) & \(\ldots\) & \(K_{\vlambda_1}^N\left(\vz_0, d\vz_1^{(N)}\right)\) \\
   \(t=2\) & \(K_{\vlambda_2}^{N + 1}\left(\vz_0, d\vz_2^{(1)}\right)\) & \(K_{\vlambda_2}^{N + 2}\left(\vz_0, d\vz_2^{(2)}\right)\) & \(K_{\vlambda_2}^{N + 3}\left(\vz_0, d\vz_2^{(3)}\right)\) & \(\ldots\) & \(K_{\vlambda_2}^{2\,N}\left(\vz_0, d\vz_2^{(N)}\right)\) \\
   \(\vdots\) & & & \(\vdots\) & & \\
   \(t=k\) & \(K_{\vlambda_k}^{\left(k-1\right)\,N + 1}\left(\vz_0, d\vz_k^{(1)}\right)\) & \(K_{\vlambda_k}^{\left(k-1\right)\,N + 2}\left(\vz_0, d\vz_k^{(2)}\right)\) & \(K_{\vlambda_k}^{\left(k-1\right)\,N + 3}\left(\vz_0, d\vz_k^{(3)}\right)\) & \(\ldots\) & \(K_{\vlambda_k}^{\left(k-1\right)\,N + N}\left(\vz_0, d\vz_k^{(N)}\right)\) \\
  \end{tabular}
  \egroup
  \end{center}
  }
  where \(K_{\vlambda}\left(\vz, \cdot\right)\) is an IMH kernel.
  Therefore, the \(n\)-step transition kernel for the vector of the Markov-chain samples \(\veta = \vz^{(1:N)}\) is represented as
  \begin{align*}
  P_{\vlambda}^k\left(\veta, d\veta^{\prime}\right)
  = 
  K_{\vlambda}^{N\,\left(k-1\right) + 1}\left(\vz_1, d\vz^{\prime}_1\right)
  \,
  K_{\vlambda}^{N\,\left(k-1\right) + 2}\left(\vz_2, d\vz^{\prime}_2\right)
  \cdot
  \ldots 
  \cdot
  K_{\vlambda}^{N\,\left(k-1\right) + N}\left(\vz_N, d\vz^{\prime}_N\right).
  \end{align*}

  Now, the convergence in total variation \(d_{\mathrm{TV}}\left(\cdot, \cdot\right)\) can be shown to decrease geometrically as
  \begin{alignat}{2}
    &\DTV{P_{\vlambda}^{k}\left(\veta, \cdot\right)}{\Pi}
    \nonumber
    \\
    &\quad=
    \sup_{A}
    \abs{
      \Pi\left(A\right)
      -
      P^{k}\left(\veta, A\right)
    }
    \nonumber
    \\
    &\quad\leq
    \sup_{A}
    \Bigg|
    \int_{A}
      \pi\left(d\vz^{\prime\;(1)}\right) \times \ldots \times \pi\left(d\vz^{\prime\;(N)}\right)
    \nonumber
    &&\quad\text{\textit{Definition of \(d_{\text{TV}}\)}}
      \\
      &\qquad\qquad\qquad-
      K^{(k-1)\,N\,+1}_{\vlambda}\left(\vz^{(1)}, d\vz^{\prime\;(1)}\right) \times \ldots \times K^{k\,N}_{\vlambda}\left(\vz^{(N)}, d\vz^{\prime\;(N)}\right)
    \,\Bigg|
    \nonumber
    \\
    &\quad\leq
    \sup_{A}
    \sum_{n=1}^N
    \abs{
    \int_{A}
      \pi\left(d\vz^{(n)}\right) - K^{(k-1)\,N + n}_{\vlambda}\left(\vz^{(n)}, d\vz^{\prime\;(n)}\right) 
    }
    &&\quad\text{\textit{\cref{thm:product_measure_bound}}}
    \nonumber
    \\
    &\quad=
    \sum_{n=1}^N
    \DTV{K^{(k-1)\,N + n}_{\vlambda}\left(\vz^{(n)}, \cdot\right)}{\pi}
    &&\quad\text{\textit{Definition of \(d_{\text{TV}}\)}}
    \nonumber
    \\
    &\quad\leq
    \sum_{n=1}^N
    r^{(k-1)\,N + n}
    &&\quad\text{\textit{\cref{eq:imh_ergodicity}}}
    \nonumber
    %\label{eq:used_ergodicity}
    \\
    &\quad=
    r^{k\,N}
    \,
    r^{-N}
    \,
    \frac{r - r^{N+1}}{1 - r}
    \nonumber
    \\
    &\quad=
    \frac{r \, \left(1 - r^N\right)}{r^N \left(1 - r\right)}
    \,
    {\left( r^{N} \right)}^k.
    \nonumber
  \end{alignat}
  Although the constant depends on \(r\) and \(N\), the kernel \(P\) is geometrically ergodic and converges \(N\) times faster than the base kernel \(K\).

  \paragraph{\textbf{Bound on the Gradient Variance}}
  To analyze the variance of the gradient, we require detailed information about the \(n\)-step marginal transition kernel, which is unavailable for most MCMC kernels.
  Fortunately, specifically for the IMH kernel,~\citet{Smith96exacttransition} have shown that the \(n\)-step marginal IMH kernel is given as~\cref{eq:imh_exact_kernel}.

  Furthermore, by \cref{thm:second_moment_bound}, the second moment of the gradient is bounded as
  \begin{alignat}{2}
    \E{ {\lVert \rvvg_{t,\text{JSA}} \rVert}_2^2 \,\middle|\, \mathcal{F}_{t-1} } 
    &=
    \V{\rvvg_{t,\text{JSA}} \mid \mathcal{F}_{t-1}} + {\mathsf{Bias}\left[\rvvg_{t,\text{JSA}} \mid \mathcal{F}_{t-1}\right]}^2 + 2\, \mathsf{Bias}\left[\rvvg_{t,\text{JSA}} \mid \mathcal{F}_{t-1}\right] \norm{\vmu}_2 + \norm{\vmu}^2_2
    \nonumber
    \\
    &\leq
    \V{ \rvvg_{t,\text{JSA}} \mid \mathcal{F}_{t-1}} + {\mathsf{Bias}\left[\rvvg_{t,\text{JSA}} \mid \mathcal{F}_{t-1}\right]}^2 + 2\,L\,\mathsf{Bias}\left[\rvvg_{t,\text{JSA}} \mid \mathcal{F}_{t-1}\right] + \norm{\vmu}^2_2,
    \nonumber
  \end{alignat}
  where \(\vmu = \mathbb{E}_{\pi}\vs\left(\vlambda; \rvvz\right)\).
  As shown in \cref{thm:mcmc_bias}, the bias terms decreases in a rate of \(r^{t\,N}\).
  Therefore, 
  \begin{alignat}{2}
    \E{ {\lVert \rvvg_{t,\text{JSA}} \rVert}_2^2 \,\middle|\, \mathcal{F}_{t-1} } 
    &\leq
    \V{ \rvvg_{t,\text{JSA}} \mid \mathcal{F}_{t-1}} + \norm{\vmu}^2_2 + \mathcal{O}\left(r^{t\,N}\right).
    &&\quad\text{\textit{\cref{thm:mcmc_bias}}}
    \nonumber
  \end{alignat}
  Note that it is possible to obtain a tighter bound on the bias terms such that \(\mathcal{O}\left(r^{t\,N}/N\right)\), if we directly use \(\left(K, \rvvz\right)\) to bound the bias instead of the higher-level \(\left(P, \rvveta\right)\) abstraction.
  The extra looseness comes from the use of \cref{thm:product_measure_bound}.

  For the variance term, we show that
  \begin{alignat}{2}
    &\V{ \rvvg_{t,\mathrm{JSA}} \,\middle|\, \mathcal{F}_{t-1} }
    \nonumber
    \\
    &\;=
    \V{\frac{1}{N}\,\sum^{N}_{n=1}\vs\left(\vlambda; \rvvz^{(n)}_t\right) \,\middle|\, \mathcal{F}_{t-1} }
    \nonumber
    \\
    &\;=
      \frac{1}{N^2}\,
      \sum^{N}_{n=1}
      \V{
        \vs\left(\vlambda; \rvvz^{(n)}_t\right)
        \,\middle|\,
        \mathcal{F}_{t-1}
      }
      +
      \frac{2}{N^2}
      \sum^{N}_{n=2}
      \sum^{n-1}_{m=1}
      \Cov{
        \vs\left(\vlambda; \rvvz^{(n)}_t\right),
        \vs\left(\vlambda; \rvvz^{(m)}_t\right)
        \,\middle|\,
        \mathcal{F}_{t-1}
      }
    \nonumber
    \\
    &\;\leq
    \frac{1}{N^2}\sum^{N}_{n=1} \E{ \norm{ \vs\left(\vlambda; \rvvz^{(n)}_t\right) }^2_2 \,\middle|\, \mathcal{F}_{t-1} } + C_{\text{cov}}
    \nonumber
    \\
    &\;=
    \frac{1}{N^2}\sum^{N}_{n=1} \E{ \norm{ \vs\left(\vlambda; \rvvz^{(n)}_t\right) }^2_2 \,\middle|\, \rvvz_{t-1}^{(N)},\, \rvvlambda_{t-1} } + C_{\text{cov}}
    \nonumber
    \\
    &\;=
    \frac{1}{N^2}\sum^{N}_{n=1} \Esub{\rvvz^{(n)}_t \sim K^n_{\rvvlambda_{t-1}}\left(\vz_{t-1}, \cdot\right)}{ \norm{\vs\left(\vlambda; \rvvz^{(n)}_t\right) }^2_2 \,\middle|\,  \rvvz_{t-1}^{(N)},\, \rvvlambda_{t-1} }
    + 
    C_{\text{cov}}
    \nonumber
    \\
    &\;\leq
    \frac{1}{N^2}\sum^{N}_{n=1}
    \left[\,
      n \, {r}^{n-1}
      \Esub{\rvvz \sim q_{\text{def.}}\left(\cdot; \rvvlambda_{t-1}\right)}{ \norm{\vs\left(\vlambda; \rvvz \right)}^2_2 }
      + 
      {r}^{n} \, \norm{\vs\left(\vlambda; \rvvz_{t-1}^{(N)} \right)}^2_2
      \,\right]
      +
    C_{\text{cov}}
    &&\quad\text{\textit{\cref{thm:imh_expecation}}}
    \nonumber
    \\
    &\;\leq
    \frac{1}{N^2}\sum^{N}_{n=1}
    \left[\,
      n \, {r}^{n-1}  \, L^2
      +
      {r}^{n} \, L^2
      \,\right]
      +
    C_{\text{cov}}
    &&\quad\text{\textit{\cref{thm:bounded_score}}}
    \nonumber
    \\
    &\;=
    \frac{L^2}{N^2}\sum^{N}_{n=1}
    \left[\,
      n \, {\left(1 - \frac{1}{w^*}\right)}^{n-1}
      +
      {\left(1 - \frac{1}{w^*}\right)}^{n} 
      \,\right]
      +
    C_{\text{cov}}
    \nonumber
    \\
    &\;=
    \frac{L^2}{N^2} \,
    \left[\,
      {\left(w^*\right)}^2 + w^*
      -
      {\left(1 - \frac{1}{w^*}\right)}^N
      \left(
        {\left(w^*\right)}^2 + w^* + N\,w^*
      \right)
    \,\right]
    +
    C_{\text{cov}}
    \nonumber
    \\
    &\;=
    \frac{L^2}{N^2} \,
    \left[\,
    \frac{1}{2} N^2 + \frac{3}{2}\,N 
    + \mathcal{O}\left(1/w^*\right)
    \,\right]
    +
    C_{\text{cov}}
    \nonumber
    &&\quad\text{\textit{Laurent series expansion at \(w^* \rightarrow \infty\)}}
    \\
    &\;=
    L^2 \,
    \left[\,
    \frac{1}{2} + \frac{3}{2}\,\frac{1}{N}
    + \mathcal{O}\left(1/w^*\right)
    \,\right] + C_{\text{cov}},
    \nonumber
  \end{alignat}
  where \[
  C_\text{cov} = {\frac{2}{N^2} \sum^{N}_{n=2} \sum^{n-1}_{m = 1} \Cov{\vs\left(\vlambda; \rvvz^{(n)}_t\right), \vs\left(\vlambda; \rvvz^{(m)}_t\right) \,\middle|\, \rvvz_{t-1}^{(N)},\, \rvvlambda_{t-1} }}.\]
  The Laurent approximation becomes exact as \(w^* \rightarrow \infty\), which is useful considering \cref{thm:wstar}.
\end{proofEnd}

%%% Local Variables:
%%% TeX-master: "master"
%%% End:

%
\vspace{-1.5ex}
\paragraph{Discussion}
As shown in~\cref{thm:jsa}, JSA benefits from increasing \(N\) in terms of a faster mixing rate.
However,  under lack of convergence and model misspecification (large \(w^*\)), the variance improvement becomes marginal.
Specifically, in the large \(w^*\) regime, the variance reduction is limited by the constant \(1/2\) term.
This is true even when, ideally, the covariance between the samples is ignorable such that \(C_{\text{cov}} \approx 0\).
In practice, however, the covariance term \(C_{\text{cov}}\) will be positive, only increasing variance.
Therefore, \textit{in the large \(w^*\) regime, JSA will perform poorly, and the variance reduction by increasing \(N\) is fundamentally limited}.
%Furthermore, the inefficiency persists even under stationarity unless \(\pi\left(\cdot\right) \in \mathcal{Q}\) such that the rejection rate becomes exactly 0.

%Intuitively, the inefficiency comes from the rejections persisting even under stationarity.
%On challenging problems with a large \(w^*\), the performance of JSA will also be poor.

%
\begin{wrapfigure}[12]{r}{0.43\textwidth}
\vspace{-7ex}
\begin{minipage}[c]{0.43\textwidth}
  \begin{algorithm2e}[H]
    \DontPrintSemicolon
    \SetAlgoLined
    \KwIn{initial samples \(\vz_0^{(1)},\, \ldots,\, \vz_0^{(N)}\),\linebreak
      initial parameter \(\vlambda_0\), \linebreak
      number of iterations \(T\), \linebreak
      stepsize schedule \(\gamma_t\)
    }
    \For{\(t = 1, 2, \ldots, T\)}{
        \For{\(n = 1, 2, \ldots, N\)}{
          \(\vz^{(n)}_{t} \sim K_{\vlambda_{t-1}}(\vz^{(n)}_{t-1}, \cdot)\)\;
      }
      \( \vg\left(\vlambda\right) = -\frac{1}{N} \sum^{N}_{n=1} \vs\,(\vlambda; \vz_{t}^{(n)}) \)\;
      \( \vlambda_{t} = \vlambda_{t-1} - \gamma_t\, \vg\left(\vlambda_{t-1}\right) \)\;
    }
    \caption{pMCSA}\label{alg:pmcsa}
  \end{algorithm2e}
\end{minipage}
\end{wrapfigure}
\vspace{-1.ex}
\section{Parallel Markov Chain Score Ascent}\label{section:pmcsa}
\vspace{-1.5ex}
%\vspace{-0.12in}
Our analysis in~\cref{section:comparison} suggests that the statistical performance of MSC, MSC-RB, and JSA are heavily affected by model specification and the state of convergence through \(w^*\).
%This affects both the mixing rate \(\rho\) and the gradient variance \(G\).
Furthermore, for JSA, a large \(w^*\) abolishes our ability to counterbalance the inefficiency by increasing the computational budget \(N\).
However, \(\rho\) and \(G\) do not equally impact convergence; recent results on MCGD suggest that gradient variance is more critical than the mixing rate (see \cref{section:convergence}).
We turn to leverage this understanding to overcome the limitations of previous methods.

%\vspace{-0.1in}
\vspace{-1.ex}
\subsection{Parallel Markov Chain Score Ascent}
\vspace{-1ex}
We propose a novel scheme, \textit{parallel Markov chain score ascent} (pMCSA,~\cref{alg:pmcsa}), that embraces a slower mixing rate in order to consistently achieve an \(\mathcal{O}\left(\nicefrac{1}{N}\right)\) variance reduction, even on challenging problems with a large \(w^*\),  

%\jrg{can we motivate this more strongly? Specifically, can we say that, by tying MCSA theory to MCGD theory, we can observe that the mixing rate isn't as important, so we develop a new scheme to exploit this previously not understood property.} 

\vspace{-1.5ex}
\paragraph{Algorithm Description}
Unlike JSA that uses \(N\) \textit{sequential} Markov chain states, pMCSA operates \(N\) \textit{parallel} Markov chains.
To maintain a similar per-iteration cost with JSA, it performs only a single Markov chain transition for each chain.
Since the chains are independent, the Metropolis-Hastings rejections do not affect the variance of pMCSA.

\begin{theoremEnd}{theorem}\label{thm:pmcsa}
  pMCSA, our proposed scheme, is obtained by setting
  {%\small
  \begin{align*}
    P_{\vlambda}^k\left(\veta, d\veta^{\prime}\right)
    = 
    K_{\vlambda}^k\left(\vz^{(1)}, d\vz^{\prime\; (1)}\right)
    \,
    K_{\vlambda}^k\left(\vz^{(2)}, d\vz^{\prime\; (2)}\right)
    \cdot
    \ldots 
    \cdot
    K_{\vlambda}^k\left(\vz^{(N)}, d\vz^{\prime\; (N)}\right)
  \end{align*}
  }
  with \(\veta = \left(\vz^{(1)}, \vz^{(2)}, \ldots, \vz^{(N)}\right)\).
  Then, given~\cref{thm:bounded_weight,thm:bounded_score}, the mixing rate and the gradient variance bounds are
  {\small
  \begin{align*}
    \textstyle
    \DTV{P_{\vlambda}^k\left(\veta, \cdot\right)}{\Pi}
    \leq
    C\left(N\right)\,{r^k}
    \;\;\text{and}\;\;
    \E{ {\lVert\rvvg_{t,\mathrm{pMCSA}}\rVert}^2_2 \,\middle|\, \mathcal{F}_{t-1} }
    \leq
    L^2 \left[\; \frac{1}{N} + \frac{1}{N}\,\left(1 - \frac{1}{w^*}\right) \;\right]
    +
    \mathcal{O}\left(r^{t}\right)
    +
    {\lVert\vmu\rVert}^2_2,
  \end{align*}
  }%
  where
  \(\vmu = \mathbb{E}_{\pi}\vs\left(\vlambda;\rvvz\right)\), 
  \(w^* = \sup_{\vz} \pi\left(\vz\right) / q_{\text{def.}}\left(\vz;\vlambda\right)\) and \(C\left(N\right) > 0\) is a finite constant.
\end{theoremEnd}
\begin{proofEnd}

  Our proposed scheme, pMCSA, is described in~\cref{alg:jsa}. 
  At each iteration, our scheme performs a single MCMC transition for each of the \(N\) samples, or chains, to estimate the gradient.
  That is,
  \begin{align*}
    \rvvz^{(1)}_{t} &\mid \rvvz_{t-1}^{(1)},\, \rvvlambda_{t-1} \sim K_{\rvvlambda_{t-1}}\left(\rvvz_{t-1}^{(1)}, \cdot\right) \\
    \rvvz^{(2)}_{t} &\mid \rvvz_{t-1}^{(2)},\, \rvvlambda_{t-1}  \sim K_{\rvvlambda_{t-1}}\left(\rvvz_{t-1}^{(2)}, \cdot\right) \\
    &\qquad\qquad\vdots
    \\
    \rvvz^{(N)}_{t} &\mid \rvvz^{(N)}_{t-1},\, \rvvlambda_{t-1}  \sim K_{\rvvlambda_{t-1}}\left(\rvvz_{t-1}^{(N)}, \cdot\right)
    \\
    \rvvg_{t,\text{pMCSA}} &= -\frac{1}{N}\,\sum^{N}_{n=1} \, \vs\left(\vlambda, \rvvz^{(n)}_t\right),
  \end{align*}
  where \(K_{\rvvlambda_{t-1}}^n\) is an \(n\)-transition IMH kernel using \(q_{\text{def.}}\left(\cdot; \rvvlambda_{t-1}\right)\).

  \paragraph{Ergodicity of the Markov Chain}
  Since our kernel operates the same MCMC kernel \(K_{\vlambda}\) for each of the \(N\) parallel Markov chains, the \(n\)-step marginal kernel \(P_{\vlambda}\) can be represented as
  \begin{align*}
    P_{\vlambda}^k\left(\veta, d\veta^{\prime}\right)
    = 
    K_{\vlambda}^k\left(\vz^{(1)}, d\vz^{\prime\; (1)}\right)
    \,
    K_{\vlambda}^k\left(\vz^{(2)}, d\vz^{\prime\; (2)}\right)
    \cdot
    \ldots 
    \cdot
    K_{\vlambda}^k\left(\vz^{(N)}, d\vz^{\prime\; (N)}\right).
  \end{align*}
  Then, the convergence in total variation \(d_{\mathrm{TV}}\left(\cdot, \cdot\right)\) can be shown to decrease geometrically as
  \begin{alignat}{2}
    &\DTV{K^{k}_{\vlambda}\left(\veta, \cdot\right)}{\Pi}
    \nonumber
    \\
    &\quad=
    \sup_{A}
    \abs{
      \Pi\left(A\right)
      -
      P^{k}_{\vlambda}\left(\veta, A\right)
    }
    &&\quad\text{\textit{Definition of \(d_{\text{TV}}\)}}
    \nonumber
    \\
    &\quad\leq
    \sup_{A}
    \big|\;
    \int_{A}
      \pi\left(d\vz^{\prime}_1\right) \cdot \ldots \cdot \pi\left(d\vz^{\prime}_N\right)
    \nonumber
      \\
      &\qquad\qquad\qquad-
      K^k_{\vlambda}\left(\vz_1, d\vz^{\prime}_1\right) \cdot \ldots \cdot K^k_{\vlambda}\left(\vz_N, d\vz^{\prime}_N\right)
    \;\big|
    \nonumber
    \\
    &\quad\leq
    \sup_{A}
    \sum_{n=1}^N
    \abs{
    \int_{A}
      \pi\left(d\vz^{\prime}_k\right) - K^{k}_{\vlambda}\left(\vz_n, d\vz^{\prime}_n\right) 
    }
    &&\quad\text{\textit{\cref{thm:product_measure_bound}}}
    \nonumber
    \\
    &\quad=
    \sum_{n=1}^N
    \DTV{K^k_{\vlambda}\left(\vz_n, \cdot\right)}{\pi}
    &&\quad\text{\textit{\cref{eq:imh_ergodicity}}}
    \nonumber
    \\
    &\quad\leq
    \sum_{n=1}^N
    r^{k}
    &&\quad\text{\textit{Geometric ergodicity}}
    \nonumber
    \\
    &\quad=
    N\,r^{k}.
    \nonumber
  \end{alignat}

  \paragraph{\textbf{Bound on the Gradient Variance}}
  By \cref{thm:second_moment_bound}, the second moment of the gradient is bounded as
  \begin{alignat}{2}
    \E{ {\lVert \rvvg_{t,\text{pMCSA}} \rVert}_2^2 \,\middle|\, \mathcal{F}_{t-1} } 
    &=
    \V{\rvvg_{t,\text{pMCSA}} \mid \mathcal{F}_{t-1}} + {\mathsf{Bias}\left[\rvvg_{t,\text{pMCSA}} \mid \mathcal{F}_{t-1}\right]}^2 + 2\, \mathsf{Bias}\left[\rvvg_{t,\text{pMCSA}} \mid \mathcal{F}_{t-1}\right] \norm{\vmu}_2 + \norm{\vmu}^2_2
    \nonumber
    \\
    &\leq
    \V{ \rvvg_{t,\text{pMCSA}} \mid \mathcal{F}_{t-1}} + {\mathsf{Bias}\left[\rvvg_{t,\text{pMCSA}} \mid \mathcal{F}_{t-1}\right]}^2 + 2\,L\,\mathsf{Bias}\left[\rvvg_{t,\text{pMCSA}} \mid \mathcal{F}_{t-1}\right] + \norm{\vmu}^2_2,
    \nonumber
  \end{alignat}
  where \(\vmu = \mathbb{E}_{\pi}\vs\left(\vlambda; \rvvz\right)\).
  As shown in \cref{thm:mcmc_bias}, the bias terms decreases in a rate of \(r^{t}\).
  Therefore, 
  \begin{alignat}{2}
    \E{ {\lVert \rvvg_{t,\text{pMCSA}} \rVert}_2^2 \,\middle|\, \mathcal{F}_{t-1} } 
    &\leq
    \V{ \rvvg_{t,\text{pMCSA}} \mid \mathcal{F}_{t-1}} + \norm{\vmu}^2_2 + \mathcal{O}\left(r^{t}\right).
    &&\quad\text{\textit{\cref{thm:mcmc_bias}}}
    \nonumber
  \end{alignat}
  As noted in the proof of \cref{thm:jsa}, it is possible to obtain a tighter bound on the bias terms such that \(\mathcal{O}\left(r^{t}/N\right)\).

  The variance term is bounded as
  \begin{alignat}{2}
    &\V{ \rvvg_{t,\text{pMCSA}} \mid \mathcal{F}_{t-1}}
    \nonumber
    \\
    &\quad=
    \V{ \frac{1}{N}\sum^{N}_{n=1} \vs\left(\vlambda; \rvvz^{(n)}_t\right) \,\middle|\, \rvvz_{t-1}^{(1:N)},\, \rvvlambda_{t-1} }
    \nonumber
    \\
    &\quad=
    \frac{1}{N^2}\sum^{N}_{n=1} \V{ \vs\left(\vlambda; \rvvz^{(n)}_t\right) \,\middle|\, \rvvz_{t-1}^{(1:N)},\, \rvvlambda_{t-1} }
    &&\quad\text{\textit{\(\rvvz^{(i)}_t \,\bot\, \rvvz^{(j)}_t\) for \(i \neq j\)}}
    \nonumber
    \\
    &\quad\leq
    \frac{1}{N^2}\sum^{N}_{n=1} \E{ \norm{\vs\left(\vlambda; \rvvz^{(n)}_t\right) }^2_2 \,\middle|\, \rvvz_{t-1}^{(1:N)},\, \rvvlambda_{t-1} }_2
    \nonumber
    \\
    &\quad=
    \frac{1}{N^2}\sum^{N}_{n=1} \Esub{\rvvz^{(n)}_t \sim K_{\rvvlambda_{t-1}}\left(\rvvz_{t-1}^{(n)}, \cdot\right)}{ \norm{\vs\left(\vlambda; \rvvz^{(n)}_t\right) }^2_2 \,\middle|\,  \rvvz_{t-1}^{(1:N)},\, \rvvlambda_{t-1} }
    \nonumber
    \\
    &\quad\leq
    \frac{1}{N^2}\sum^{N}_{n=1} \left[\,
      \Esub{\rvvz \sim q_{\text{def.}}\left(\cdot;\rvvlambda_{t-1}\right)}{ \norm{\vs\left(\vlambda; \rvvz \right)}^2_2 }
      +
      {r} \, \norm{\vs\left(\vlambda; \rvvz_{t-1}^{(n)} \right)}^2_2
      \,\right]
    &&\quad\text{\textit{\cref{thm:imh_expecation}}}
    \nonumber
    \\
    &\quad\leq
    \frac{1}{N^2}\sum^{N}_{n=1} \left[\,
        L^2 + {r} \, L^2 \,\right]
    &&\quad\text{\textit{\cref{thm:bounded_score}}}
    \nonumber
    \\
    &\quad=
    \frac{L^2}{N^2} \sum^{N}_{n=1} \left[\,
      1 + {r} \,\right]
    \nonumber
    \\
    &\quad=
    L^2 \left[\, \frac{1}{N} + \frac{1}{N}\,{\left(1 - \frac{1}{w^*}\right)} \,\right].
    \nonumber
  \end{alignat}
\end{proofEnd}

%%% Local Variables:
%%% TeX-master: "master"
%%% End:

\vspace{-1.5ex}
\paragraph{Discussion}
Unlike JSA and MSC, the variance reduction rate of pMCSA is independent of \(w^*\).
Therefore, it should perform significantly better on challenging practical problems.
If we consider the rate of~\citet{duchi_ergodic_2012}, the combined rate is constant with respect to \(N\) since it cancels out.
In practice, however, we observe that increasing \(N\) accelerates convergence quite dramatically.
Therefore, the mixing rate independent convergence rates by~\citet{doan_finitetime_2020, doan_convergence_2020} appears to better reflect practical performance.
This is because
\begin{enumerate*}[label=\textbf{(\roman*)}]
  \item the mixing rate \(\rho\) is a conservative \textit{global} bound and 
  \item the mixing rate will improve naturally as MCSA converges.
\end{enumerate*}

\begin{wraptable}[11]{r}{0.55\textwidth}
  \vspace{-5ex}
  \centering
  
% Second version of table, with booktabs.
%\begin{table}
%\centering
\caption{Computational Costs}\label{table:cost}
\setlength{\tabcolsep}{1.5pt}
  \begin{threeparttable}
\begin{tabular}{lccccc}\toprule
& \multicolumn{3}{c}{\footnotesize Applying \(P_{\vlambda}\)} & \multicolumn{2}{c}{\footnotesize Estimating \(\vg\)} \\
\cmidrule(lr){2-4}\cmidrule(lr){5-6}
  & {\footnotesize\(p\left( \vz, \vx \right)\)}
  & {\footnotesize\(q\left(\vz; \vlambda\right)\)}
  & {\footnotesize\(q\left(\vz; \vlambda\right)\)}
  & {\footnotesize\(p\left( \vz, \vx \right)\)}
  & {\footnotesize\( q\left(\vz; \vlambda\right)\)}
  \\
  & {\footnotesize\# Eval.  }
  & {\footnotesize\# Eval.  }
  & {\footnotesize\# Samples}
  & {\footnotesize\# Grad.  }
  & {\footnotesize\# Grad.  }
\\\midrule
{%\footnotesize
ELBO
}
& \(0\)
& \(0\)
& \(N\)
& \(N\)
& \(N\)
\\\arrayrulecolor{black!30}\midrule
{%\footnotesize
MSC
}
& \(N-1\)
& \(N\)
& \(N-1\)
& \(0\)
& \(1\)
\\
{%\footnotesize
MSC-RB
}
& \(N-1\)
& \(N\)
& \(N-1\)
& \(0\)
& \(N\)
\\
{%\footnotesize
JSA
}
& \(N\)
& \(N+1\)
& \(N\)
& \(0\)
& \(N\)
\\
{%\footnotesize
{\textbf{pMCSA}}
}
& \(N\)
& \(2 \, N\)
& \(N\)
& \(0\)
& \(N\)
\\\bottomrule
\end{tabular}
%% \begin{tablenotes}[flushleft]
%%     \item[]{\footnotesize\textbf{\textit{Note}:} We assume that the variables are cached as much as possible.}
%%   \end{tablenotes}
\end{threeparttable}
%\end{table}

\end{wraptable}
\vspace{-1.5ex}
\subsection{Computational Cost Comparison}
\vspace{-1.5ex}
The four schemes using the CIS and IMH kernels have different costs depending on \(N\) as organized in~\cref{table:cost}.

\vspace{-1.5ex}
\paragraph{Cost of Sampling Proposals}
For the CIS kernel used by MSC, \(N\) controls the number of internal proposals sampled from \(q\,(\cdot; \vlambda)\).
For JSA and pMCSA, the IMH kernel only uses a single sample from \(q\,(\cdot; \vlambda)\), but applies the kernel \(N\) times.
On the other hand, pMCSA needs twice more evaluations of \(q\,(\cdot; \vlambda)\).
However, this added cost is minimal since it is dominated by that of evaluating \(p\,(\vz,\vx)\).

\vspace{-1.5ex}
\paragraph{Cost of Estimating the Score}
When estimating the score, MSC computes \(\nabla_{\vlambda} \log q\,(\vz; \vlambda)\) only once, while JSA and our proposed scheme compute it \(N\) times.
However, MSC-RB also computes the score \(N\) times.
Lastly, notice that MCSA methods do not differentiate through the likelihood \(p\,(\vz,\vx)\), unlike ELBO maximization, making its per-iteration cost significantly cheaper.

%%% Local Variables:
%%% TeX-master: "master"
%%% End:

\vspace{-1ex}
\section{Evaluations}\label{section:eval}
\vspace{-1.5ex}
\subsection{Experimental Setup}
\vspace{-1.5ex}
%\vspace{-0.05in}
\paragraph{Implementation}
For the realistic experiments, we implemented\footnote{Available at \url{https://github.com/Red-Portal/KLpqVI.jl}} MCSA methods on top of the Turing~\citep{ge2018t} probabilistic programming framework.
For the variational family, we use diagonal multivariate Gaussians (mean-field family) with the support transformation of~\citet{JMLR:v18:16-107}.
We use the ADAM optimizer by~\citet{kingma_adam_2015} with a stepsize of 0.01 in all experiments.
The budget is set to \(N=10\) for all experiments unless specified.

\begin{figure*}[t]
  \vspace{-3ex}
  \centering
  \includegraphics[scale=1.0]{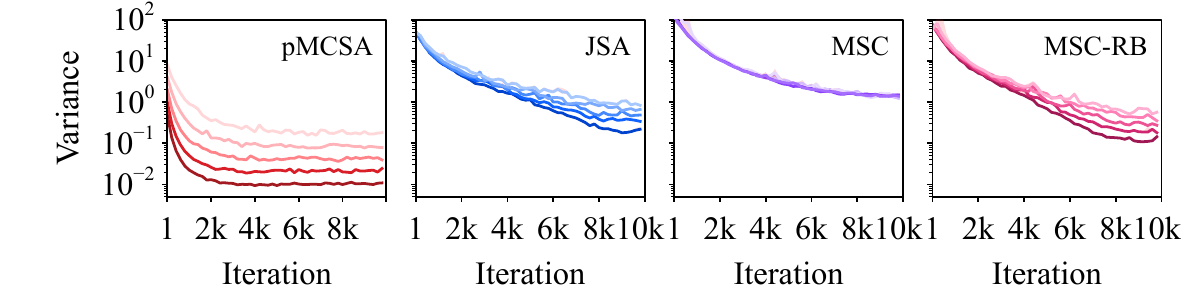}
  \vspace{-0.5ex}
  \caption{\textbf{Gradient variance versus iteration and computational budget (\(N\)).
      pMCSA not only achieves the least gradient variance, but its variance also scales better with \(N\).
    }
    The colors range from light (\(N=2^3\)) to dark (\(N=2^7\)) representing the computational budgets \(N \in [2^3, 2^4, 2^5, 2^6, 2^7]\).
    The target distribution is a 50-D multivariate Gaussian with \(\nu = 500\).
    The error bands are the 80\% quantiles obtained from 8 replications.
  }\label{fig:gaussian}
  \vspace{-1.5ex}
\end{figure*}
\begin{figure*}[t]
  \centering
  \includegraphics[scale=1.0]{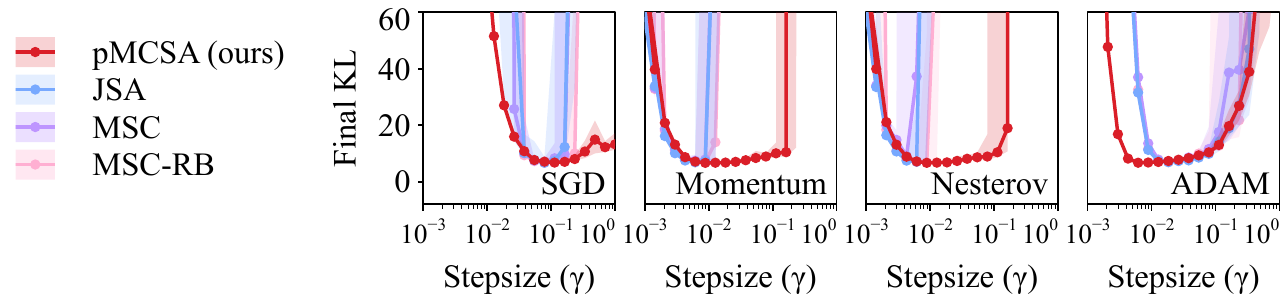}
  %\vspace{-0.05in}
  \caption{\textbf{Optimizer stepsize (\(\gamma\)) versus final KL.
      pMCSA is the least sensitive to optimizer hyperparameters and results in stable convergence.}
      The final KL is obtained at the \(10^4\)th iteration.
      The target distribution is a 100-D Gaussian with \(\nu = 500\).
      The error bands are the 80\% quantiles, while the solid lines are the median of 20 replications.
  }\label{fig:stepsize}
  \vspace{-2ex}
\end{figure*}

\begin{table*}[t]
  \vspace{-2ex}
  \centering
  \caption{Test Log Predictive Density on \textbf{Bayesian Neural Network Regression}}\label{table:bnn}
  %\vspace{-0.05in}
  \setlength{\tabcolsep}{2pt}
  \begin{threeparttable}
  \begin{tabular}{lrrrcccccc}
    \toprule
    & \multicolumn{1}{c}{\multirow{2}{*}{\(D_{\mathcal{\vlambda}}\)}} & \multicolumn{1}{c}{\multirow{2}{*}{\(D_{\mathcal{\vx}}\)}} &  \multicolumn{1}{c}{\multirow{2}{*}{\(N_{\text{train}}\)}} & \multicolumn{2}{c}{\multirow{1}{*}{ELBO}} & \multicolumn{4}{c}{MCSA Variants} \\\cmidrule(lr){5-6}\cmidrule(lr){7-10}
  & & & & {\footnotesize\(N=1\)} & {\footnotesize\(N=10\)} & \multicolumn{1}{c}{\multirow{1}{*}{\footnotesize{\textbf{pMCSA{\scriptsize\,(ours)}}}}} & \multicolumn{1}{c}{\multirow{1}{*}{\footnotesize{JSA}}} & \multicolumn{1}{c}{\multirow{1}{*}{\footnotesize{MSC}}} & \multicolumn{1}{c}{\multirow{1}{*}{\footnotesize{MSC-RB}}}\\
    \midrule
    \textsf{yacht} & 403 & 6 & 277 & {\bf-2.45 {\scriptsize{\(\pm 0.01\)}}} & {\bf-2.44 {\scriptsize{\(\pm 0.01\)}}} & {\bf-2.49 {\scriptsize{\(\pm 0.01\)}}} & {-3.00 {\scriptsize{\(\pm 0.05\)}}} & {-2.98 {\scriptsize{\(\pm 0.04\)}}} & {-2.98 {\scriptsize{\(\pm 0.04\)}}}\\
    \textsf{concrete} & 503 & 8 & 927 & {-3.25 {\scriptsize{\(\pm 0.01\)}}} & {\bf-3.24 {\scriptsize{\(\pm 0.01\)}}} & {\bf-3.20 {\scriptsize{\(\pm 0.01\)}}} & {-3.33 {\scriptsize{\(\pm 0.02\)}}} & {-3.32 {\scriptsize{\(\pm 0.02\)}}} & {-3.33 {\scriptsize{\(\pm 0.02\)}}}\\
    \textsf{airfoil} & 353 & 6 & 1352 & {-2.53 {\scriptsize{\(\pm 0.02\)}}} & {-2.56 {\scriptsize{\(\pm 0.02\)}}} & {\bf-2.27 {\scriptsize{\(\pm 0.02\)}}} & {-2.51 {\scriptsize{\(\pm 0.02\)}}} & {-2.53 {\scriptsize{\(\pm 0.01\)}}} & {-2.51 {\scriptsize{\(\pm 0.01\)}}}\\
    \textsf{energy} & 503 & 9 & 691 & {-2.42 {\scriptsize{\(\pm 0.02\)}}} & {-2.40 {\scriptsize{\(\pm 0.02\)}}} & {\bf-1.92 {\scriptsize{\(\pm 0.03\)}}} & {-2.38 {\scriptsize{\(\pm 0.02\)}}} & {-2.37 {\scriptsize{\(\pm 0.02\)}}} & {-2.36 {\scriptsize{\(\pm 0.02\)}}}\\
    \textsf{wine} & 653 & 12 & 1439 & {\bf-0.96 {\scriptsize{\(\pm 0.01\)}}} & {\bf-0.96 {\scriptsize{\(\pm 0.01\)}}} & {\bf-0.95 {\scriptsize{\(\pm 0.01\)}}} & {\bf-0.97 {\scriptsize{\(\pm 0.01\)}}} & {\bf-0.97 {\scriptsize{\(\pm 0.01\)}}} & {\bf-0.97 {\scriptsize{\(\pm 0.01\)}}} \\
    \textsf{boston} & 753 & 14 & 455 & {\bf-2.72 {\scriptsize{\(\pm 0.03\)}}} & {\bf-2.70 {\scriptsize{\(\pm 0.03\)}}} & {\bf-2.69 {\scriptsize{\(\pm 0.02\)}}} & {-2.82 {\scriptsize{\(\pm 0.02\)}}} & {-2.80 {\scriptsize{\(\pm 0.03\)}}} & {-2.78 {\scriptsize{\(\pm 0.02\)}}}\\
    \textsf{sml} & 1203 & 23 & 3723 & {-1.32 {\scriptsize{\(\pm 0.01\)}}} & {\bf-1.25 {\scriptsize{\(\pm 0.02\)}}} & {\bf-1.22 {\scriptsize{\(\pm 0.01\)}}} & {-1.72 {\scriptsize{\(\pm 0.01\)}}} & {-1.97 {\scriptsize{\(\pm 0.02\)}}} & {-1.95 {\scriptsize{\(\pm 0.02\)}}} \\
    \textsf{gas} & 6503 & 129 & 2308 & {\bf-0.06 {\scriptsize{\(\pm 0.01\)}}} & {\bf 0.13 {\scriptsize{\(\pm 0.03\)}}} & {-0.09 {\scriptsize{\(\pm 0.02\)}}} & {-0.47 {\scriptsize{\(\pm 0.03\)}}} & {-0.47 {\scriptsize{\(\pm 0.04\)}}} & {-0.50 {\scriptsize{\(\pm 0.03\)}}}\\
    \bottomrule
 \end{tabular}
  \begin{tablenotes}
    \item[1] {\footnotesize \(D_{\vlambda}\): Dimentionality of \(\vlambda\), \(D_{\vx}\): Number of features, \(N_{\text{train}}\): Number of training data points.}
    \item[2] {\footnotesize \(\pm\) denotes the 95\% bootstrap confidence intervals obtained from 20 replications.}
    \item[3] {\footnotesize Bolded numbers don't have enough evidence to be distinguished from the best performing method under a .05 significance threshold (Friedman test with Nemenyi post-hoc test,~\citealt{JMLR:v7:demsar06a})}.
  \end{tablenotes}
  \end{threeparttable}
  \vspace{-2ex}
\end{table*}

%%% Local Variables:
%%% TeX-master: "master"
%%% End:

\vspace{-1.5ex}
\paragraph{Baselines}
We compare
\begin{enumerate*}[label=\textbf{(\roman*)}]
  \item \textbf{pMCSA} (ours,~\cref{section:pmcsa}),
  \item \textbf{JSA}~\citep{pmlr-v124-ou20a},
  \item \textbf{MSC}~\citep{NEURIPS2020_b2070693},
  \item MSC with with Rao-Blackwellization (\textbf{MSC-RB},~\citealt{NEURIPS2020_b2070693}), and
  %\item adaptive importance sampling (IS) with the self-normalized IS estimator (\textbf{SNIS},~\citealt{robert_monte_2004}), and
  \item evidence lower-bound maximization (\textbf{ELBO},~\citealt{pmlr-v33-ranganath14, JMLR:v18:16-107}) with the path derivative estimator~\citep{NIPS2017_e91068ff}.
\end{enumerate*}
\pagebreak

\vspace{-1.5ex}
\subsection{Simulations}\label{section:simulation}
\vspace{-1.5ex}
\paragraph{Setup}
First, we verify our theoretical analysis on multivariate Gaussians with full-rank covariances sampled from Wishart distribution with \(\nu\) degrees of freedom (values of \(\nu\) are in the figure captions).
This problem is challenging since an IMH (used by pMCSA, JSA) or CIS (used by MSC, MSC-RB) kernel with a diagonal Gaussian proposal will mix slowly due to a large \(w^*\).
% Since the exact KL divergence between the target and the varitional approximation is available, accurate evaluation is possible.

\vspace{-1.5ex}
\paragraph{Gradient Variance}
We evaluate our theoretical analysis of the gradient variance.
The variance is estimated from \(512\) independent Markov chains using the parameters generated by the main MCSA procedure.
The estimated variances are shown in \cref{fig:gaussian}.
We make the following observations:
\begin{enumerate*}[label=\textbf{(\roman*)}]
  \item pMCSA has the lowest variance overall, and it consistently benefits from increasing \(N\).
  \item MSC does not benefit from increasing \(N\) whatsoever.
  \item MSC-RB does not benefit much from increasing \(N\) until the 
\(\chi^2\) divergence between \(\pi\) and \(q\left(\cdot; \vlambda\right)\) has become small.
  \item JSA does not benefit from increasing \(N\) until \(q\left(\cdot; \vlambda\right)\) has sufficiently converged (when \(w^*\) has become small).
\end{enumerate*}
These results confirm our theoretical analyses in~\cref{section:comparison,section:pmcsa}.

\vspace{-1.5ex}
\paragraph{Robustness Against Optimizers}
Since the convergence of most sophisticated SGD optimizers has yet to be established for MCGD, we empirically investigate their effectiveness.
The results using SGD~\citep{robbins_stochastic_1951, bottou_optimization_2018a}, Momentum~\citep{polyak_methods_1964}, Nesterov~\citep{nesterov_method_1983}, ADAM~\citep{kingma_adam_2015}, and varying stepsizes are shown in~\cref{fig:stepsize}.
Clearly, pMCSA successfully converges for the broadest variety of optimizer settings.
Overall, most MCSA methods seem to be the most stable with ADAM, which points out that establishing the convergence of ADAM for MCGD will be a promising direction for future works.

%% \subsection{Hierarchical Logistic Regression}\label{section:logistic}
%% \vspace{-0.05in}
%% \paragraph{Experimental Setup}
%% We now perform logistic regression with the \texttt{Pima Indians} diabetes (\(\vz \in \mathbb{R}^{11}\),~\citealt{smith_using_1988}), \texttt{German credit} (\(\vz \in \mathbb{R}^{27}\)), and \texttt{heart disease} (\(\vz \in \mathbb{R}^{16}\),~\citealt{detrano_international_1989}) datasets obtained from the UCI repository~\citep{Dua:2019}.
%% 10\% of the data points were randomly selected in each of the 100 repetitions as test data.

% \input{gp_table.tex}

%% %

%%   \vspace{-0.1in}
%% \paragraph{Inclusive KL v.s. Exclusive KL}
%% While both ELBO and par.-IMH showed similar numerical performance, they chose different optimization paths in the parameter space.
%% This is shown in~\cref{fig:logistic}.
%% While the test accuracy suggests that ELBO converges quickly around \(t=2000\) (\cref{fig:german_acc}), in terms of uncertainty estimate, it takes much longer to converge (\cref{fig:german_lpd}).
%% This shows that inclusive KL minimization chooses a path that has better density coverage as expected.

  \vspace{-0.1in}
\subsection{Bayesian Neural Network Regression}\label{section:bnn}
  \vspace{-0.07in}
\looseness=-1
\begin{wrapfigure}[12]{r}{0.5\textwidth}
  \vspace{-5.ex}
  \centering
  \includegraphics[scale=0.9]{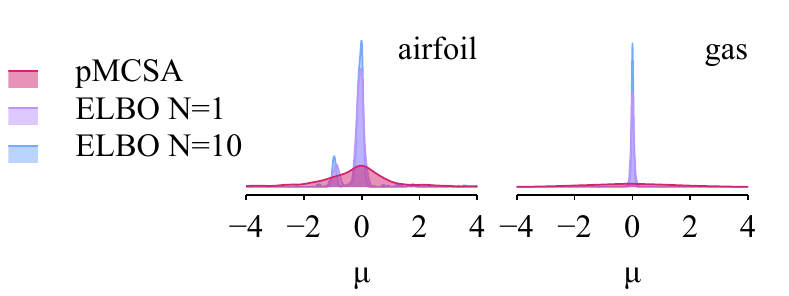}
  \vspace{-2ex}
  \caption{\textbf{
      Distribution of the variational posterior mean of the BNN weights.
      pMCSA results in much less pruning.
    }
    The density was estimated with a Gaussian kernel and the bandwidth was selected with Silverman's rule.
  }\label{fig:pruning}
\end{wrapfigure}
\paragraph{Setup}
For realistic experiments, we train Bayesian neural networks (BNN, \citealt{neal_bayesian_1996}) for regression.
We use datasets from the UCI repository~\citep{Dua:2019} with 90\% random train-test splits and run for \(T=5\cdot10^4\) iterations.
We use the model and forward propagation method of~\citet{pmlr-v37-hernandez-lobatoc15} with a \(50\)-unit hidden layer (see \cref{section:model_bnn}).
% \looseness=-1

\vspace{-1.5ex}
\paragraph{Results}
The results are shown in~\cref{table:bnn}.
pMCSA achieves the best performance compared to all other MCSA methods.
Also, its overall performance is comparable to exclusive KL minimization methods (ELBO) unlike other MCSA methods.
Furthermore, on \textsf{airfoil} and \textsf{energy}, pMCSA improves over ELBO by 0.29 \texttt{nat} and 0.48 \texttt{nat}.
Even on \textsf{gas} where pMCSA did not beat ELBO, its performance is comparable, and it dominates all other MCSA methods by roughly 0.4 \texttt{nat}.
Additional experimental results, including root mean-square error (RMSE) results and plots with respect to the wall clock time, can be found in~\cref{section:bnn_additional}

\vspace{-1.5ex}
\paragraph{Weight Pruning}
When using VI, BNNs have been known to underfit data, which~\citet{mackay_local_2001,pmlr-v70-hoffman17a,trippe_overpruning_2017} associated with ``weight pruning.'' 
That is, the variational posterior converges to the zero-mean prior.
Furthermore,~\citet{coker_wide_2022,huix_variational_2022} have shown that this is guaranteed to happen under certain conditions.
However, these results are strictly based on exclusive KL minimization.
On the other hand,~\cref{fig:pruning} shows that pMCSA does not suffer from weight pruning, which suggests that pruning is an artifact of using the exclusive KL.
Additional plots are shown in \cref{fig:pruning_additional} (\cref{section:bnn_additional}).
%

%\begin{wraptable}{r}{0.4\textwidth}
\begin{table*}[t]
  \centering
  \vspace{-2ex}
  \caption{Test Log Predictive Density on \textbf{Robust Gaussian Process Regression}}\label{table:gp}
  \vspace{-1ex}
  %\vspace{-0.05in}
  \setlength{\tabcolsep}{4pt}
  \begin{threeparttable}
  \begin{tabular}{lrrrccccc}
    \toprule
    & \multicolumn{1}{c}{\multirow{2}{*}{\footnotesize\(D_{\vlambda}\)}} & \multicolumn{1}{c}{\multirow{2}{*}{\(D_{\vx}\)}} &  \multicolumn{1}{c}{\multirow{2}{*}{\(N_{\text{train}}\)}} & \multicolumn{1}{c}{\multirow{1}{*}{ELBO}} & \multicolumn{4}{c}{MCSA Variants} \\\cmidrule(lr){5-5}\cmidrule(lr){6-9}
   & & & & \(N=1\) & \multicolumn{1}{c}{\multirow{1}{*}{\footnotesize\textbf{pMCSA{\scriptsize\,(ours)}}}} & \multicolumn{1}{c}{\multirow{1}{*}{\footnotesize{JSA}}} & \multicolumn{1}{c}{\multirow{1}{*}{\footnotesize{MSC}}} & \multicolumn{1}{c}{\multirow{1}{*}{\footnotesize{MSC-RB}}}\\
    \midrule
    \textsf{yacht} & 287 & 6 & 277 & {-3.63 {\scriptsize{\(\pm 0.02\)}}} & {\bf-3.31 {\scriptsize{\(\pm 0.04\)}}} & {\bf-3.29 {\scriptsize{\(\pm 0.05\)}}} & {\bf-3.25 {\scriptsize{\(\pm 0.04\)}}} & {\bf-3.27 {\scriptsize{\(\pm 0.05\)}}} \\
    \textsf{airfoil} & 353 & 6 & 1352 & {-3.14 {\scriptsize{\(\pm 0.01\)}}} & {\bf-2.63 {\scriptsize{\(\pm 0.01\)}}} & {-2.83 {\scriptsize{\(\pm 0.04\)}}} & {-2.77 {\scriptsize{\(\pm 0.02\)}}} & {\bf-2.73 {\scriptsize{\(\pm 0.02\)}}}\\
    \textsf{boston} & 472 & 13 & 455 & {\bf-2.98 {\scriptsize{\(\pm 0.01\)}}} & {\bf-2.96 {\scriptsize{\(\pm 0.02\)}}} & {\bf-3.00 {\scriptsize{\(\pm 0.03\)}}} & {\bf-3.00 {\scriptsize{\(\pm 0.03\)}}} & {\bf-2.96 {\scriptsize{\(\pm 0.03\)}}} \\
    \textsf{energy} & 703 & 8 & 691 & {-2.75 {\scriptsize{\(\pm 0.01\)}}} & {\bf-2.58 {\scriptsize{\(\pm 0.03\)}}} & {-2.78 {\scriptsize{\(\pm 0.04\)}}} & {-2.70 {\scriptsize{\(\pm 0.04\)}}} & {-2.72 {\scriptsize{\(\pm 0.05\)}}} \\
    \textsf{concrete} & 939 & 8 & 927 & {-3.68 {\scriptsize{\(\pm 0.01\)}}} & {\bf-3.49 {\scriptsize{\(\pm 0.01\)}}} & {-3.69 {\scriptsize{\(\pm 0.02\)}}} & {\bf-3.59 {\scriptsize{\(\pm 0.04\)}}} & {\bf-3.57 {\scriptsize{\(\pm 0.02\)}}} \\
    \textsf{wine} & 1454 & 11 & 1439 & {-1.02 {\scriptsize{\(\pm 0.01\)}}} & {\bf-0.94 {\scriptsize{\(\pm 0.02\)}}} & {-1.04 {\scriptsize{\(\pm 0.01\)}}} & {-1.00 {\scriptsize{\(\pm 0.02\)}}} & {-0.99 {\scriptsize{\(\pm 0.02\)}}} \\
    \textsf{gas} & 2440 & 128 & 2308 & {\bf{0.18} {\scriptsize{\(\pm 0.02\)}}} & {\bf-0.86 {\scriptsize{\(\pm 0.02\)}}} & {-1.10 {\scriptsize{\(\pm 0.03\)}}} & {-1.10 {\scriptsize{\(\pm 0.04\)}}} & {-1.06 {\scriptsize{\(\pm 0.02\)}}} 
    \\\bottomrule
  \end{tabular}
  \begin{tablenotes}
    \item[1] {\footnotesize \(D_{\vlambda}\): Dimentionality of \(\vlambda\), \(D_{\vx}\): Number of features, \(N_{\text{train}}\): Number of training data points.}
    \item[2] {\footnotesize \(\pm\) denotes the 95\% bootstrap confidence intervals obtained from 20 replications.}
    \item[3] {\footnotesize Bolded numbers don't have enough evidence to be distinguished from the best performing method under a .05 significance threshold (Friedman test with Nemenyi post-hoc test,~\citealt{JMLR:v7:demsar06a})}.
  \end{tablenotes}
  \end{threeparttable}
  \vspace{-2ex}
\end{table*}
%\end{wraptable}

%%% Local Variables:
%%% TeX-master: "master"
%%% End:

\vspace{-1.5ex}
\subsection{Robust Gaussian Process Regression}\label{section:bgp}
\vspace{-1.5ex}
\paragraph{Setup}
We train Gaussian processes (GP) with a Student-T likelihood for robust regression.
We use datasets from the UCI repository~\citep{Dua:2019} with 90\% random train-test splits.
We use the Mat\'ern 5/2 covariance kernel with automatic relevance determination~\citep{neal_bayesian_1996} (see \cref{section:model_rgp}).
We run all methods with \(T=2\cdot10^4\) iterations.
For prediction, we use the mode of \(q\left(\cdot; \vlambda\right)\) for the hyperparameters and marginalize the latent function over \(q\left(\cdot; \vlambda\right)\)~\citep{rasmussen_gaussian_2006}.
We consider ELBO with only \(N=1\) since differentiating through the likelihood makes its per-iteration cost comparable to MCSA methods with \(N=10\).

\vspace{-1.8ex}
\paragraph{Results}
The results are shown in~\cref{table:gp}.
Except for \textsf{gas}, pMCSA achieves better predictive densities than all other methods.
This suggests that, overall, the exclusive KL may be less effective in terms of uncertainty quantification for GP posteriors.
While ELBO achieves the best performance on \textsf{gas}, among MCSA methods, pMCSA dominates.
Our encouraging regression results suggest that incorporating methods such as inducing points~\citep{NIPS2005_4491777b} into MCSA may lead to an important new class of GP models.
Additional experimental results, including RMSE results and plots with respect to the wall clock time, can be found in~\cref{section:gp_additional}.
Note that in~\cref{section:gp_additional}, MCSA methods appear worse in terms of RMSE compared to the exclusive KL since the inclusive KL puts less probability volume around the posterior mode.

%%% Local Variables:
%%% TeX-master: "master"
%%% End:

\vspace{-1.5ex}
\section{Related Works}\label{section:related}
\vspace{-1.5ex}
\paragraph{Inclusive KL minimization}
Our MCSA framework generalizes MSC~\citep{NEURIPS2020_b2070693} and JSA~\citet{pmlr-v124-ou20a}, which are inclusive KL minimization based on SGD and Markov chains.
Similar to MCSA is the method of~\citet{li_approximate_2017}.
However, the convergence of this method is not guaranteed since it uses short Markov chains, disqualifying for MCSA.
Other methods based on biased gradients have been proposed by~\citet{DBLP:journals/corr/BornscheinB14,le_revisiting_2020}, but these are specific for deep generative models.
On a different note,~\citet{pmlr-v161-jerfel21a} use boosting instead of SGD to minimize the inclusive KL, which gradually builds a complex variational approximation from a simple variational family.

\vspace{-1.5ex}
\paragraph{Beyond the KL Divergence}
Discovering alternative divergences for VI has been an active research area.
For example, the \(\chi^2\)~\citep{NIPS2017_35464c84}, \(f\)~\citep{NEURIPS2018_1cd138d0, NEURIPS2020_c928d86f}, \(\alpha\)~\citep{NIPS2016_7750ca35, regli_alphabeta_2018, pmlr-v48-hernandez-lobatob16}, reguarlized importance ratio~\citep{bamler_perturbative_2017} divergences have been studied for VI.
However, for gradient estimation these methods involve the importance ratio \(w\left(\vz\right) = \pi\left(\vz\right)/q\left(\vz\right)\), which leads to significant variance and low signal-to-noise ratio under model misspecification \citep{bamler_perturbative_2017, geffner2021empirical, pmlr-v139-geffner21a}.
In contrast, under stationarity, the variance of pMCSA is \({\sigma^2}/{N}\) (\(\sigma\) is the variance of the score over the posterior) regardless of model misspecification.
Meanwhile,~\citet{pmlr-v37-salimans15,pmlr-v97-ruiz19a,NEURIPS2021_05f971b5,NEURIPS2021_a1a609f1} construct implicit divergences formed by MCMC.
With the exception of \citet{pmlr-v97-ruiz19a}, most of these approaches aim to maximize auxiliary representations of the classic ELBO.
Therefore, their property is likely to be similar to the exclusive KL.

\vspace{-1.5ex}
\paragraph{Adaptive MCMC and MCSA}
As pointed out by~\citet{pmlr-v124-ou20a}, using \(q\left(\cdot; \vlambda\right)\) within the MCMC kernel makes MCSA structurally equivalent to adaptive MCMC.
In particular,~\citet{10.1007/s11222-008-9110-y, garthwaite_adaptive_2016, pmlr-v151-brofos22a, gabrie_adaptive_2022} discuss the use of stochastic approximation in adaptive MCMC.
Also,~\citet{andrieu_ergodicity_2006, keith_adaptive_2008, holden_adaptive_2009, giordani_adaptive_2010, pmlr-v151-brofos22a, habib2018auxiliary, neklyudov_metropolishastings_2019} specifically discuss adapting the propsosal of IMH kernels, and some of them use KL divergence minimization.
These methods focus on showing ergodicity the samples (\(\veta_t\) in our context) not the convergence of the variational approximation \(q\left(\cdot; \vlambda\right)\).
In this work, we focused on the convergence of \(q\left(\cdot; \vlambda\right)\), which could advance the adaptive MCMC side of the story.

%And for HMC,~\citet{zhang_variational_2018, pmlr-v139-campbell21a} have proposed to use score matching, ELBO maximization, and Stein discrepancy minimization.

%%% Local Variables:
%%% TeX-master: "master"
%%% End:

\vspace{-1.5ex}
\section{Discussions}\label{section:discussion}
\vspace{-1.5ex}
This paper presented a new theoretical framework for analyzing inclusive KL divergence minimization methods based on running SGD with Markov chains.
Furthermore, we proposed pMCSA, a new MCSA method that enjoys substantially low variance.
We have shown that this theoretical improvement translates into better empirical performance.

\vspace{-1.5ex}
\paragraph{Limitations}
Our work has three main limitations.
Firstly, since our work aims to understand existing MCSA methods, it inherits their current limitations.
For example, minibatch subsampling is challenging for models with non-factorizable likelihoods~\citep{NEURIPS2020_b2070693}.
Secondly, our theoretical analysis in~\cref{section:comparison} requires~\cref{thm:bounded_score}, which is strong, but required to connect with MCGD.
An important future direction would be to relax the assumptions needed by MCGD.
Lastly, our MCSA framework does not include models with parameterized posteriors such as variational autoencoders.

\vspace{-1.5ex}
\paragraph{Parameterized Posteriors}
On problems with parameterized posteriors, the target posterior moves around.
Therefore, quickly chasing the moving posterior with fast converging MCMC kernels is as important as achieving low variance.
Because of this, trading bias and variance is less straightforward compared to the ``static'' setting we consider.
Furthermore, the usage of expensive MCMC kernels could be beneficial as suggested by \citet{zhang_transport_2022}.

\vspace{-1.5ex}
\paragraph{Towards Alternative Divergences}
In \cref{section:eval}, we have shown that minimizing the \textit{inclusive} KL is competitive against  minimizing the \textit{exclusive} KL on general Bayesian inference problems.
Although~\citet{dhaka_challenges_2021} have shown that the inclusive KL fails on high-dimensional problems, this is only the case under the presence of strong correlations.
Before entirely ditching the inclusive KL, it is essential to ask, ``how correlated are posteriors really in practice?''
Furthermore, the true performance of alternative divergences is often masked by the limitations of the inference procedure~\citep{geffner2021empirical, pmlr-v139-geffner21a}.
Given that pMCSA significantly advances the best-known performance of inclusive KL minimization, it is possible that similar improvements could be extracted from other divergences.
To conclude, our results motivate further development of better inference algorithms for alternative divergence measures.

%An interesting direction for future research would be to invstigate whether such result 
%This is against the conclusions of~\citet{dhaka_challenges_2021} that exclusive VI should outperform inclusive VI in high-dimensions. 

%%% Local Variables:
%%% TeX-master: "master"
%%% End:

%\section*{Broader Impact}

\vspace{-2ex}
\begin{ack}
\vspace{-2ex}
  We thank Hongseok Yang for pointing us to relevant related work, Guanyang Wang for insightful discussions about the independent Metropolis-Hastings algorithm, Geon Park and Kwanghee Choi for constructive comments that enriched this paper, Christian A. Naesseth for comments about Rao-Blackwellized MSC, and Gabriel V. Cardoso for comments about BR-SNIS.
  We also acknowledge the Department of Computer Science and Engineering of Sogang University for providing computational resources.

  K. Kim was supported in part by the EPSRC through the Big Hypotheses grant [EP/R018537/1].
  J. R. Gardner was supported by NSF award [IIS-2145644].
  H. Kim was supported by the Basic Science Research Program through the National Research Foundation of Korea (NRF) funded by the Ministry of Science and ICT under Grant [NRF-2021R1A2C1095435].
\end{ack}
%\newpage
\bibliography{references}
\newpage

%%%%%%%%%%%%%%%%%%%%%%%%%%%%%%%%%%%%%%%%%%%%%%%%%%%%%%%%%%%%
\section*{Checklist}

%%% BEGIN INSTRUCTIONS %%%
%% The checklist follows the references.  Please
%% read the checklist guidelines carefully for information on how to answer these
%% questions.  For each question, change the default \answerTODO{} to \answerYes{},
%% \answerNo{}, or \answerNA{}.  You are strongly encouraged to include a {\bf
%% justification to your answer}, either by referencing the appropriate section of
%% your paper or providing a brief inline description.  For example:
%% \begin{itemize}
%%   \item Did you include the license to the code and datasets? \answerYes{See Section~\ref{gen_inst}.}
%%   \item Did you include the license to the code and datasets? \answerNo{The code and the data are proprietary.}
%%   \item Did you include the license to the code and datasets? \answerNA{}
%% \end{itemize}
%% Please do not modify the questions and only use the provided macros for your
%% answers.  Note that the Checklist section does not count towards the page
%% limit.  In your paper, please delete this instructions block and only keep the
%% Checklist section heading above along with the questions/answers below.
%%% END INSTRUCTIONS %%%

\begin{enumerate}

\item For all authors...
\begin{enumerate}
\item Do the main claims made in the abstract and introduction accurately reflect the paper's contributions and scope?
  \answerYes{} \\
  \textit{See the contribution summary in \cref{section:intro}}.
\item Did you describe the limitations of your work?
  \answerYes{} \\
  \textit{See \cref{section:discussion}}.
\item Did you discuss any potential negative societal impacts of your work?
  \answerNo{}
\item Have you read the ethics review guidelines and ensured that your paper conforms to them?
  \answerYes{}
\end{enumerate}

\item If you are including theoretical results...
  \begin{enumerate}
  \item Did you state the full set of assumptions of all theoretical results?
    \answerYes{} \\
    \textit{The key assumptions are stated in each proof.}
  \item Did you include complete proofs of all theoretical results?
    \answerYes{} \\
    \textit{See \cref{section:proofs}.}
\end{enumerate}

\item If you ran experiments...
\begin{enumerate}
\item Did you include the code, data, and instructions needed to reproduce the main experimental results (either in the supplemental material or as a URL)?
  \answerYes{} \\
  \textit{It is included in the supplementary material}.
\item Did you specify all the training details (e.g., data splits, hyperparameters, how they were chosen)?
  \answerYes{} \\
  \textit{See \cref{section:eval}. Additional details can be found in the code in the supplementary material.}
\item Did you report error bars (e.g., with respect to the random seed after running experiments multiple times)?
  \answerYes{} \\
  \textit{See the text in \cref{section:eval}.}
\item Did you include the total amount of compute and the type of resources used (e.g., type of GPUs, internal cluster, or cloud provider)?
  \answerYes{} \\
  \textit{See \cref{section:resources}.}
\end{enumerate}

\item If you are using existing assets (e.g., code, data, models) or curating/releasing new assets...
\begin{enumerate}
  \item If your work uses existing assets, did you cite the creators?
    \answerNo{}
  \item Did you mention the license of the assets?
    \answerNo{}
  \item Did you include any new assets either in the supplemental material or as a URL?
    \answerNA{}
  \item Did you discuss whether and how consent was obtained from people whose data you're using/curating?
    \answerNA{}
  \item Did you discuss whether the data you are using/curating contains personally identifiable information or offensive content?
    \answerNA{}
\end{enumerate}

\item If you used crowdsourcing or conducted research with human subjects...
\begin{enumerate}
  \item Did you include the full text of instructions given to participants and screenshots, if applicable?
    \answerNA{}
  \item Did you describe any potential participant risks, with links to Institutional Review Board (IRB) approvals, if applicable?
    \answerNA{}
  \item Did you include the estimated hourly wage paid to participants and the total amount spent on participant compensation?
    \answerNA{}
\end{enumerate}

\end{enumerate}

\newpage
\appendix

% \begin{center}
%   \LARGE\bf
%   Markov Chain Score Ascent: \\
%   A Unifying Framework \textit{of} \\
%   Variational Inference \textit{with} Score Ascent \\
%   \textit{Appendix}
% \end{center}
%% \begin{center}
%%   \bf
%%   Markov-Chain Monte Carlo Score Estimators for \\ Variational Inference with Score Climbing
%%   \textit{Appendix}
%% \end{center}

%\section{Additional Experimental Results}
\section{Computational Resources}\label{section:resources}
\begin{table}[h]
  \centering
  \caption{Computational Resources for Bayesian Neural Network Regression}
  \begin{threeparttable}
  \begin{tabular}{ll}
    \toprule
    \multicolumn{1}{c}{\textbf{Type}}
    & \multicolumn{1}{c}{\textbf{Model and Specifications}}
    \\ \midrule
    System Topology & 4 nodes with 20 logical threads each  \\
    Processor       & Intel Xeon Xeon E5--2640 v4, 2.2 GHz (maximum 3.1 GHz) \\
    Cache           & 32 kB L1, 256 kB L2, and 25 MB L3 \\
    Memory          & 64GB RAM
    \\ \bottomrule
  \end{tabular}
  \end{threeparttable}
\end{table}
\begin{table}[h]
  \centering
  \caption{Computational Resources for Robust Gaussian Process Regression}
  \begin{threeparttable}
  \begin{tabular}{ll}
    \toprule
    \multicolumn{1}{c}{\textbf{Type}}
    & \multicolumn{1}{c}{\textbf{Model and Specifications}}
    \\ \midrule
    System Topology & 1 node with 16 logical threads \\
    Processor       & AMD EPYC 7262, 3.2 GHz (maximum 3.4 GHz) \\
    Accelerator     & NVIDIA Titan RTX, 1.3 GHZ, 24GB RAM \\
    Cache           & 256 kB L1, 4MiB L2, and 128MiB L3 \\
    Memory          & 126GB RAM
    \\ \bottomrule
  \end{tabular}
  \end{threeparttable}
\end{table}

\section{Pseudocodes}\label{section:pseudocode}

\subsection{Markov Chain Monte Carlo Kernels}

\begin{center}
\begin{minipage}[c]{0.63\textwidth}
  \begin{algorithm2e}[H]
    \DontPrintSemicolon
    \SetAlgoLined
    \KwIn{previous sample \(\vz_{t-1}\),\linebreak
      previous parameter \(\vlambda_{t-1}\),\linebreak
      number of proposals \(N\)
    }
    \(\vz^{(0)} = \vz_{t-1}\) \;
    \(\vz^{(i)} \sim q_{\text{def.}}\left(\vz;\vlambda_{t-1}\right)\quad\) for \(i = 1, 2,\ldots, N\) \;
    \(\widetilde{w}(\vz^{(i)}) = p(\vz^{(i)},\vx) \,/\, q_{\text{def.}}\left(\vz^{(i)}; \vlambda_{t-1}\right)\quad\) for \(i = 0, 1,\ldots, N\)\;
    \(\overline{w}^{(i)} = \frac{\widetilde{w}(\vz^{(i)})}{ \sum^{N}_{i=0} \widetilde{w}(\vz^{(i)}) }\quad\) for \(i = 0, 1,\ldots, N\)\;
    \(\vz_{t} \sim \mathrm{Multinomial}(\overline{w}^{(0)}, \overline{w}^{(1)}, \ldots, \overline{w}^{(N)}) \)\;
    \caption{Conditional Importance Sampling Kernel}
  \end{algorithm2e}
\end{minipage}
\end{center}

\begin{center}
\begin{minipage}[c]{0.63\textwidth}
  \begin{algorithm2e}[H]
    \DontPrintSemicolon
    \SetAlgoLined
    \KwIn{previous sample \(\vz_{t-1}\),\linebreak
      previous parameter \(\vlambda_{t-1}\),
    }
    \;
    \(\vz^* \sim q_{\text{def.}}\left(\vz; \vlambda_{t-1}\right)\)\;
    \(\widetilde{w}(\vz) = p(\vz,\vx)/q_{\text{def.}}\left(\vz; \vlambda_{t-1}\right) \)\;
    \(\alpha = \min\left( \widetilde{w}\,(\vz^*)/\widetilde{w}\,(\vz_{t-1}), 1\right)\)\;
    \(u \sim \mathrm{Uniform}(0, 1) \)\;
    \eIf{u < \(\alpha\)}
        {
          %\(\vz_t \leftarrow \vz^*\)
          \(\vz_t = \vz^*\)
        }
        {
          %\(\vz_t \leftarrow \vz_{t-1}\)
          \(\vz_t = \vz_{t-1}\)
        }
        \caption{Independent Metropolis-Hastings Kernel}
  \end{algorithm2e}
\end{minipage}
\end{center}

\subsection{Markov Chain Score Ascent Algorithms}

\begin{center}
\begin{minipage}[c]{0.63\textwidth}
  \centering
  \begin{algorithm2e}[H]
    \DontPrintSemicolon
    \SetAlgoLined
    \KwIn{Initial sample \(\vz_0\),\linebreak
      initial parameter \(\vlambda_0\),\linebreak
      number of iterations \(T\),\linebreak
      stepsize schedule \(\gamma_t\)}
    \For{\textcolor{black}{\(t = 1, 2, \ldots, T\)}}{
      \( \vz_{t} \sim K_{\vlambda_{t-1}}(\vz_{t-1}, \cdot) \)\;
      \( \vg\left(\vlambda\right) = -\vs\,(\vlambda; \vz_{t}) \)\;
      \( \vlambda_{t} = \vlambda_{t-1} - \gamma_t\, \vg\left(\vlambda_{t-1}\right) \)\;
    }
    \caption{Markovian Score Climbing}\label{alg:msc}
  \end{algorithm2e}
\end{minipage}
\end{center}

  \begin{center}
  \begin{minipage}[c]{0.63\textwidth}
    \begin{algorithm2e}[H]
      \DontPrintSemicolon
      \SetAlgoLined
      \KwIn{Initial sample \(\vz_0^{(N)}\),\linebreak
        initial parameter \(\vlambda_0\),\linebreak
        number of iterations \(T\),\linebreak
        stepsize schedule \(\gamma_t\)
      }
      \For{\(t = 1, 2, \ldots, T\)}{
        \(\vz_{t}^{(0)} = \vz_{t-1}^{(N)}\)\;
        \For{\(n = 1, 2, \ldots, N\)}{
          \(\vz_{t}^{(n)} \sim K_{\vlambda_{t-1}}(\vz_{t}^{(n-1)}, \cdot)\)\;
        }
        \( \vg\left(\vlambda\right) = -\frac{1}{N} \sum^{N}_{n=1} \vs\,(\vlambda; \vz_{t}^{(n)}) \)\;
        \( \vlambda_{t} = \vlambda_{t-1} - \gamma_t \, \vg\left(\vlambda_{t-1}\right) \,
        \)\;
      }
      \caption{Joint Stochastic Approximation}\label{alg:jsa}
    \end{algorithm2e}
  \end{minipage}
  \end{center}

%%% Local Variables:
%%% TeX-master: "master"
%%% End:

\newpage

\section{Probabilistic Models Used in the Experiments}
\subsection{Bayesian Neural Network Regression}\label{section:model_bnn}
We use the BNN model of~\citet{pmlr-v37-hernandez-lobatoc15} defined as
\begin{align*}
  \lambda^{-1} &\sim \mathsf{inverse\text{-}gamma}\left(\alpha=6, \beta=6\right) \\
  \gamma^{-1}  &\sim \mathsf{inverse\text{-}gamma}\left(\alpha=6, \beta=6\right) \\
  \mW_1       &\sim \mathcal{N}\left(\mathbf{0}, \lambda^{-1} \mI \right) \\
  \vz         &= \mathsf{ReLU}\left(\mW_1 \vx_i\right) \\
  \mW_2       &\sim \mathcal{N}\left(\mathbf{0}, \lambda^{-1} \mI \right) \\
  \widehat{y} &= \mathsf{ReLU}\left(\mW_2 \vz\right) \\
  y_i         &\sim \mathcal{N}\left(\widehat{y}, \gamma^{-1}\right),
\end{align*}
where \(\vx_i\) and \(y_i\) are the feature vector and target value of the \(i\)th datapoint.
Given the variational distribution of \(\lambda^{-1}, \gamma^{-1}, \mW_1, \mW_2\), we use the same posterior predictive approximation of~\citet{pmlr-v37-hernandez-lobatoc15}.
We apply z-standardization (whitening) to the features \(\vx_i\) and the target values \(y_i\), and unwhiten the predictive distribution.

\subsection{Robust Gaussian Process Logistic Regression}\label{section:model_rgp}
We perform robust Gaussian process regression by using a student-t prior with a latent Gaussian process prior.
The model is defined as
\begin{align*}
   \log \sigma_f &\sim \mathcal{N}(0, 4) \\
   \log \epsilon &\sim \mathcal{N}(0, 4) \\
   \log \ell_i   &\sim \mathcal{N}(0, 0.2) \\
   f   &\sim \mathcal{GP}\left(\mathbf{0}, \mSigma_{\sigma_f, \mathbf{\ell}} + \left(\delta + \epsilon^2\right)\,\mI \right) \\
   \nu &\sim \mathsf{gamma}\left(\alpha=4, \beta=1/10\right) \\
   \log \sigma_y &\sim \mathcal{N}(0, 4) \\
   y_i &\sim \mathsf{student\text{-}t}\left(  f\left( \vx_i \right), \sigma_y, \nu  \right).
\end{align*}
The covariance \(\mSigma\) is computed using a kernel \(k\left(\cdot, \cdot\right)\) such that \({[\mSigma]}_{i,j} = k\left( \vx_{i}, \vx_{j} \right) \) where \(\vx_i\) and \(\vx_j\) are data points in the dataset.
For the kernel, we use the Matern 5/2 kernel with automatic relevance determination~\citep{neal_bayesian_1996} defined as
\begin{align*}
  &k\left(\vx, \vx' ;\; \sigma^2, \ell^2_1, \ldots, \ell^2_D \right) =
  \sigma_f \left( 1 + \sqrt{5} r + \frac{5}{3} r^2 \right) \exp\left( - \sqrt{5} r \right), \quad
  \text{where}\;\; r = \sum^{D}_{i=1} \frac{ {\left(\vx_i - \vx'_i\right)}^2 }{\ell^2_i}
\end{align*}
and \(D\) is the number of dimensions.
The jitter term \(\delta\) is used for numerical stability.
We set a small value of \(\delta = 1\times10^{-6}\).

%% \subsection{Radon Hierarchical Regression}
%% The partially pooled linear regression model used in~\cref{section:mll} is
%% \begin{align*}
%%   \sigma_{a_1} &\sim \mathrm{Gamma}\left( \alpha = 1, \beta = 0.02 \right) \\
%%   \sigma_{a_2} &\sim \mathrm{Gamma}\left( \alpha = 1, \beta = 0.02 \right) \\
%%   \sigma_{y}  &\sim \mathrm{Gamma}\left( \alpha = 1, \beta = 0.02 \right) \\
%%   \mu_{a_1}    &\sim \mathcal{N}\left( 0, 1 \right) \\
%%   \mu_{a_2}    &\sim \mathcal{N}\left( 0, 1 \right) \\
%%   a_{1,\, c}     &\sim \mathcal{N}\left( \mu_{a_1}, \sigma_{a_1}^2 \right) \\
%%   a_{2,\, c}     &\sim \mathcal{N}\left( \mu_{a_2}, \sigma_{a_2}^2 \right) \\
%%   y_i         &\sim \mathcal{N}\left( a_{1,\, c_i} + a_{2,\, c_i}\,x_i,\, \sigma_y^2 \right)
%% \end{align*}
%% where \(a_{1,\,c}\) is the intercept at the county \(c\), \(a_{2,\,c}\) is the slope at the county \(c\), \(c_i\) is the county of the \(i\)th datapoint, \(x_i\) and \(y_i\) are the floor predictor of the measurement and the measured radon level of the \(i\)the datapoint, respectively.
%% The model pools the datapoints into their respective counties, which complicates the posterior geometry~\citep{betancourt_hierarchical_2020}.

%%% Local Variables:
%%% TeX-master: "master"
%%% End:

\newpage
\section{Proofs}\label{section:proofs}
\printProofs
\newpage

\section{Additional Experimental Results}
%% \subsection{Bayesian Neural Network Regression}\label{section:gaussian_additional}
%% \begin{figure}
%%   \centering
%% %%   \includegraphics[scale=1.0]{figures/gaussian_02.pdf}
%%   \includegraphics[scale=1.0]{figures/gaussian_04.pdf}
%%   \vspace{-0.08in}
%%   \caption{\textbf{Gradient variance versus iteration and computational budget (\(N\)).
%%       pMCSA not only achieves the least gradient variance, but its variance also scales better with \(N\).
%%     }
%%     The colors range from light (\(N=2^3\)) to dark (\(N=2^7\)) representing the computational budgets \(N \in [2^3, 2^4, 2^5, 2^6, 2^7]\).
%%     The target distribution is a 100-D multivariate Gaussian with \(\nu = 500\).
%%     The error bands are the 80\% quantiles obtained from 8 replications.
%%   }
%% \end{figure}

\subsection{Bayesian Neural Network Regression}\label{section:bnn_additional}
\begin{figure}[ht]
  \centering
  \subfloat[\textsf{yacht}]{
    \includegraphics[scale=0.9]{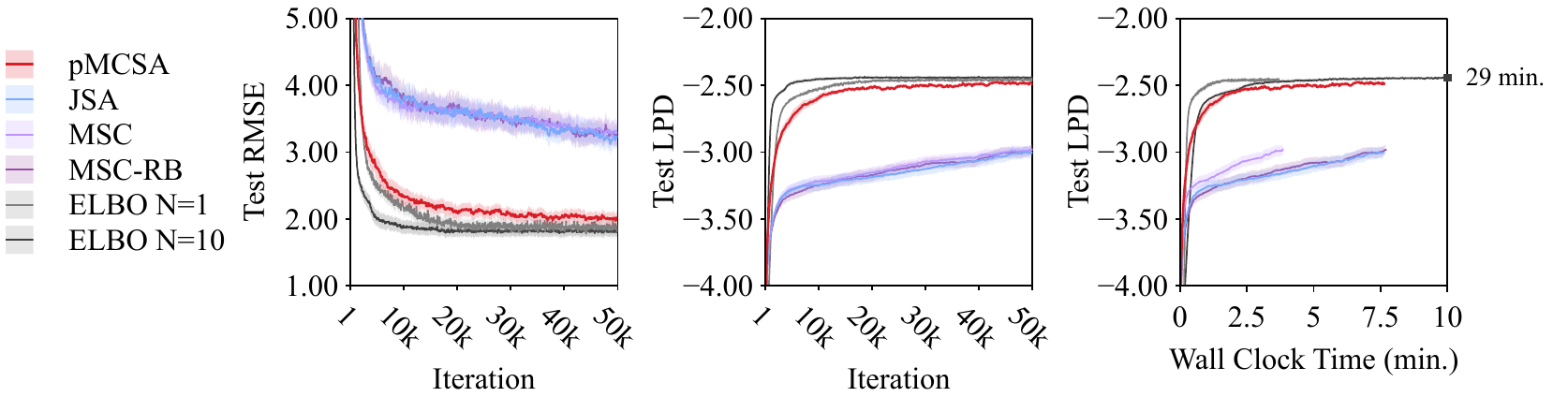}
    \vspace{-0.05in}
  }\vspace{0.05in}\\
  \subfloat[\textsf{concrete}]{
    \includegraphics[scale=0.9]{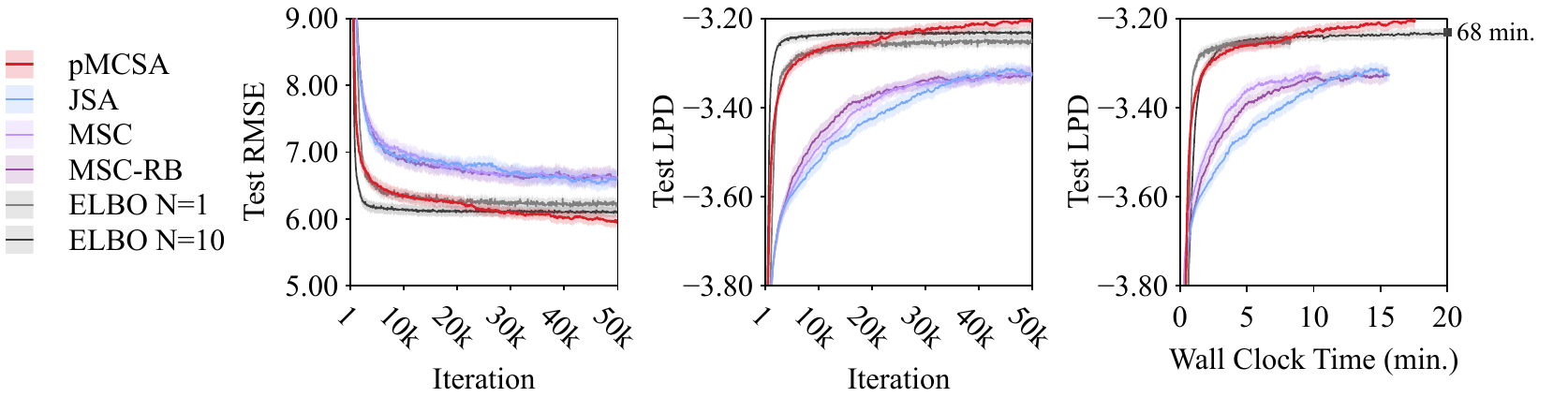}
    \vspace{-0.05in}
  }\vspace{0.05in}\\
  \subfloat[\textsf{airfoil}]{
    \includegraphics[scale=0.9]{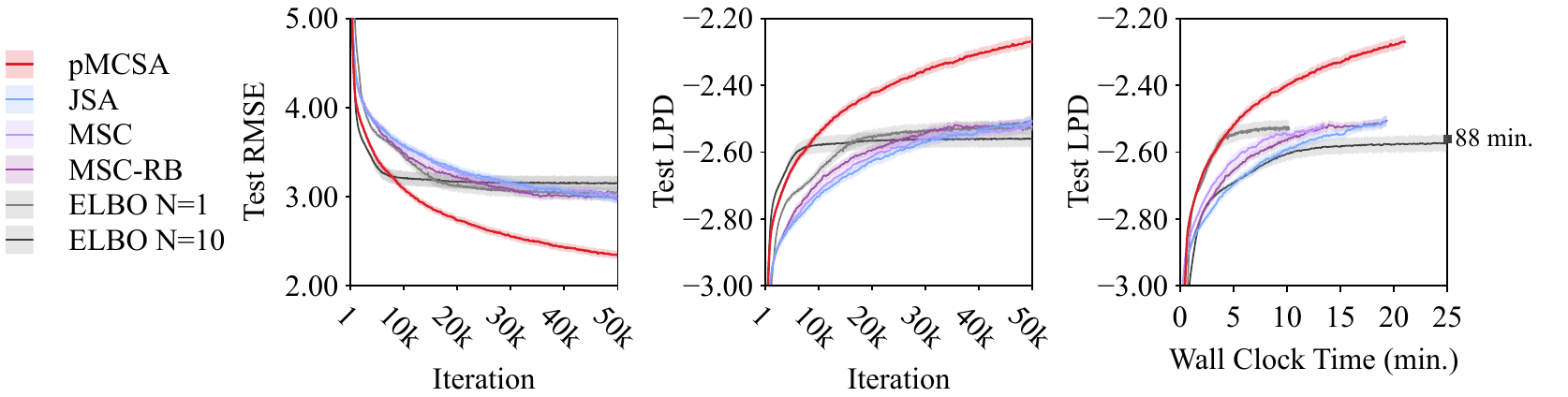}
    \vspace{-0.05in}
  }\vspace{0.05in}\\
  \subfloat[\textsf{energy}]{
    \includegraphics[scale=0.9]{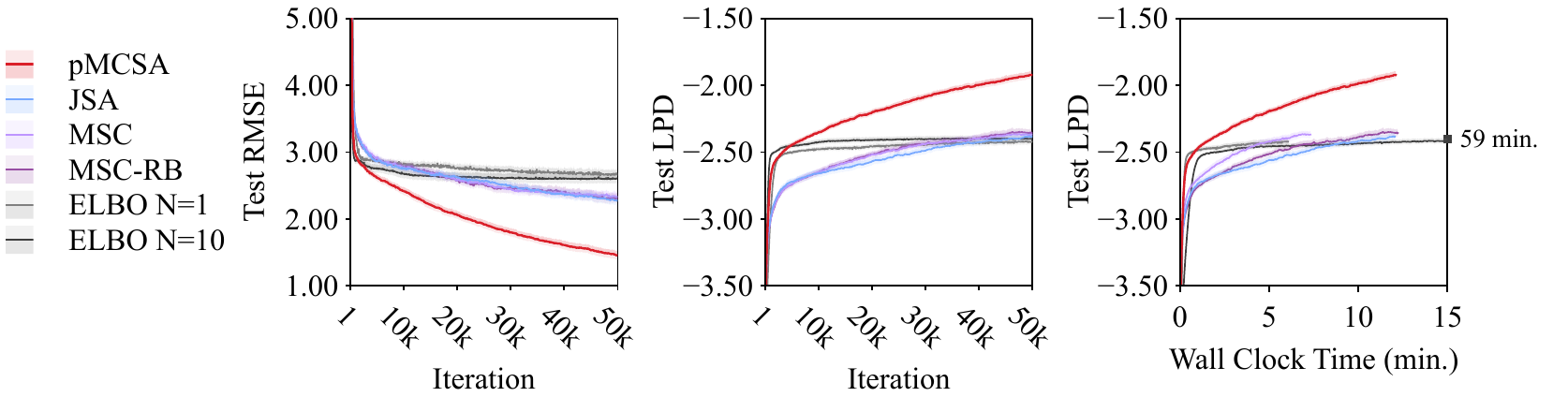}
    \vspace{-0.05in}
  }
  \caption{\textbf{Test root-mean-square error (RMSE) and test log predictive density (LPD) on Bayesian neural network regression.} 
    The grey squares (\textcolor{darkgrey}{\(\blacksquare\)}) mark the performance of ELBO \(N=10\) at the wall clock time shown next to it.
  The error bands show the 95\% bootstrap confidence intervals obtained from 20 independent 90\% train-test splits.}
\end{figure}
\newpage
\begin{figure}[h]
  \centering
  \subfloat[\textsf{wine}]{
    \includegraphics[scale=0.9]{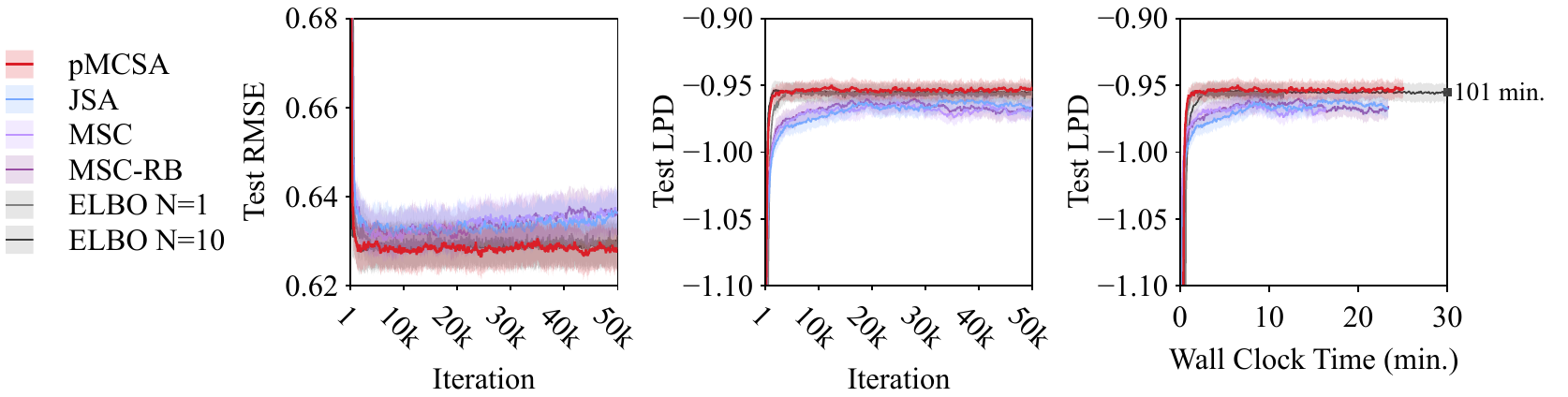}
    \vspace{-0.05in}
  }\vspace{0.05in}\\
  \subfloat[\textsf{boston}]{
    \includegraphics[scale=0.9]{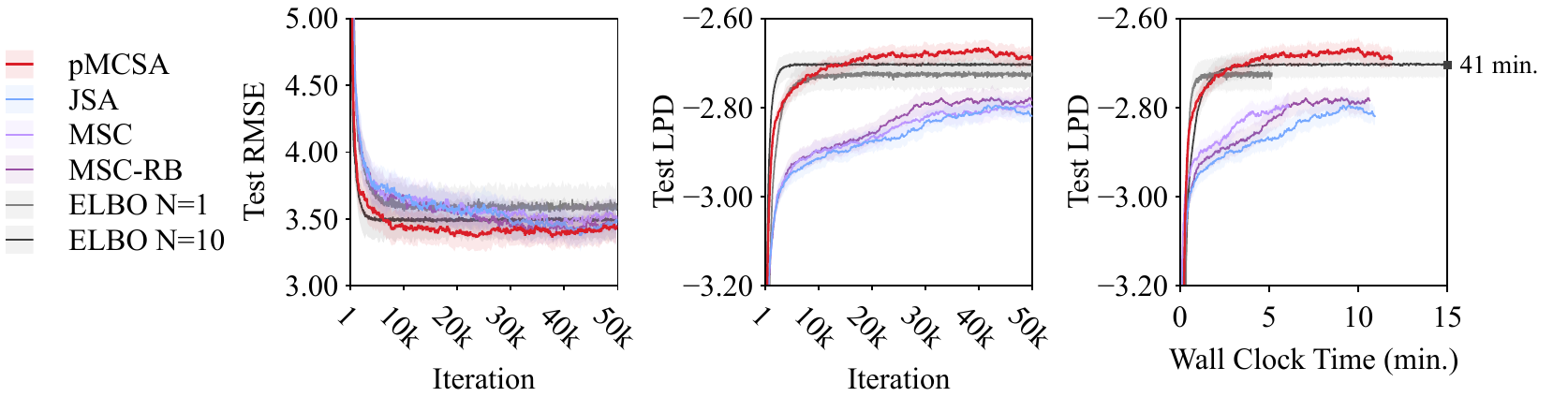}
    \vspace{-0.05in}
  }\vspace{0.05in}\\
  \subfloat[\textsf{sml}]{
    \includegraphics[scale=0.9]{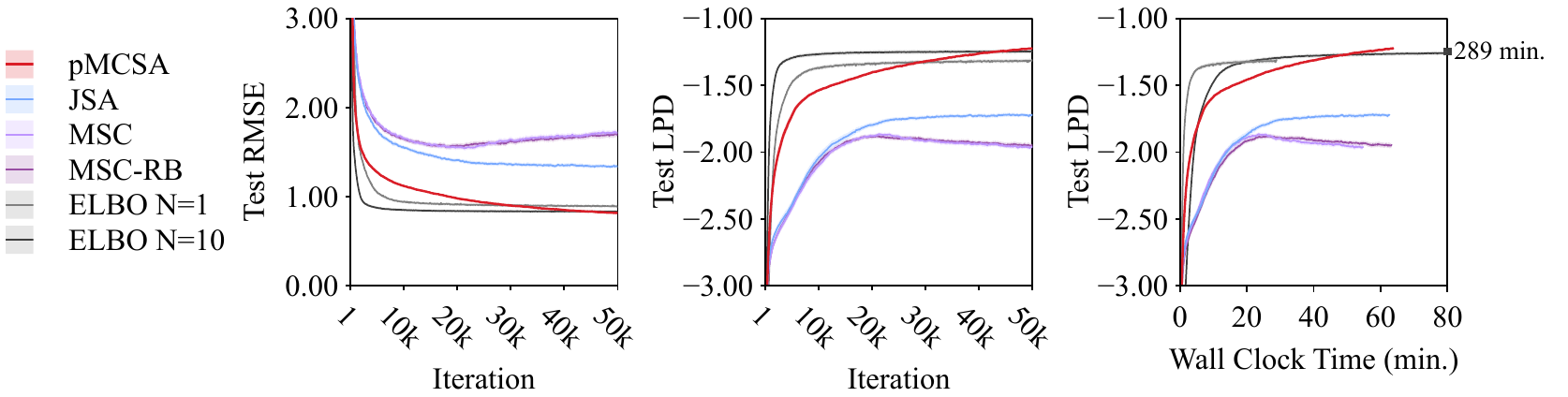}
    \vspace{-0.05in}
  }\vspace{0.05in}\\
  \subfloat[\textsf{gas}]{
    \includegraphics[scale=0.9]{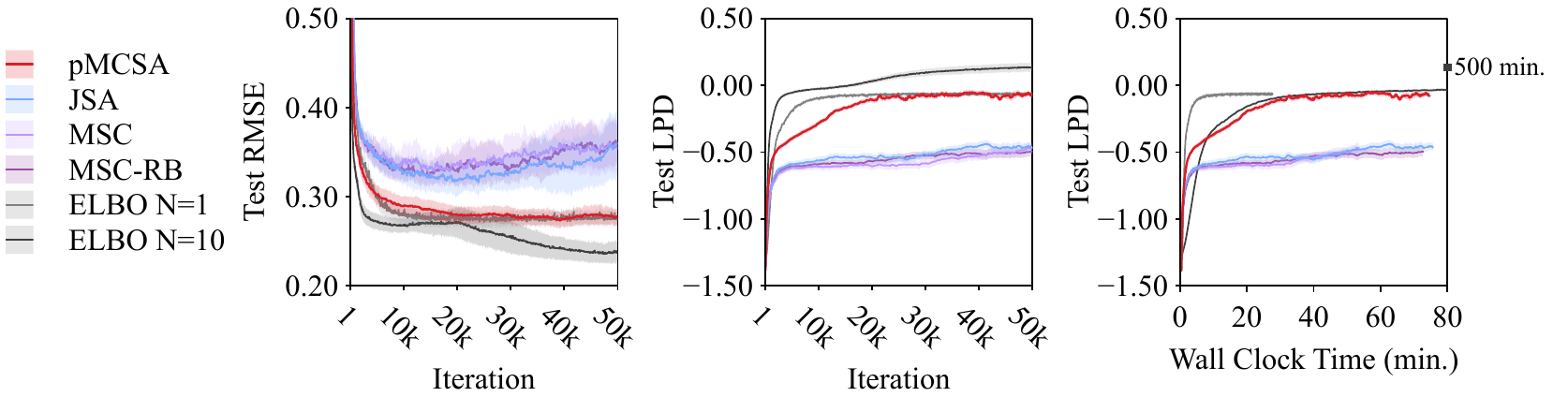}
    \vspace{-0.05in}
  }
  \caption{\textbf{(continued) Test root-mean-square error (RMSE) and test log predictive density (LPD) on Bayesian neural network regression.}
    The grey squares (\textcolor{darkgrey}{\(\blacksquare\)}) mark the performance of ELBO \(N=10\) at the wall clock time shown next to it.
  The error bands show the 95\% bootstrap confidence intervals obtained from 20 independent 90\% train-test splits.}
\end{figure}

\newpage
\begin{figure}[H]
  \vspace{-0.10in}
  \centering
  \includegraphics[scale=0.9]{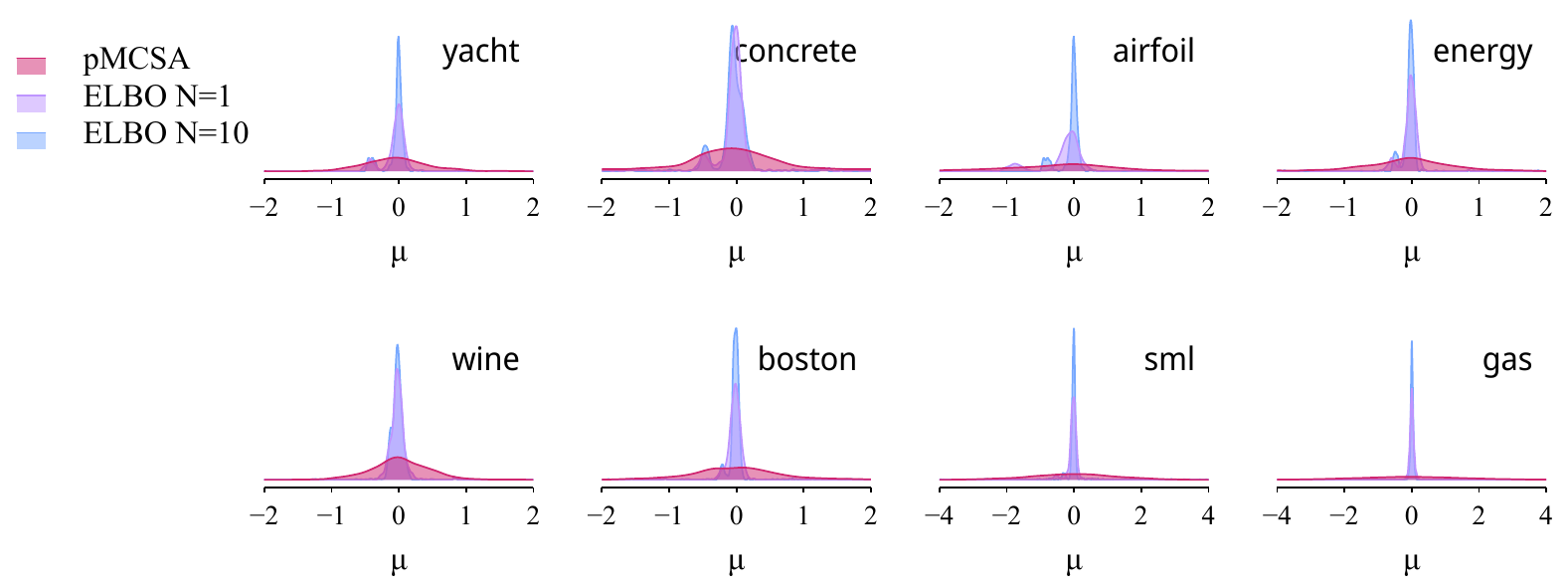}
  \vspace{-0.05in}
  \caption{\textbf{
      Distribution of the variational posterior mean of the BNN weights.
    }
    The density was estimated with a Gaussian kernel and the bandwidth was selected with Silverman's rule
  }\label{fig:pruning_additional}
\end{figure}

\subsection{Robust Gaussian Process Regression}\label{section:gp_additional}
\begin{figure}[h]
  \centering
  \subfloat[\textsf{yacht}]{
    \includegraphics[scale=0.9]{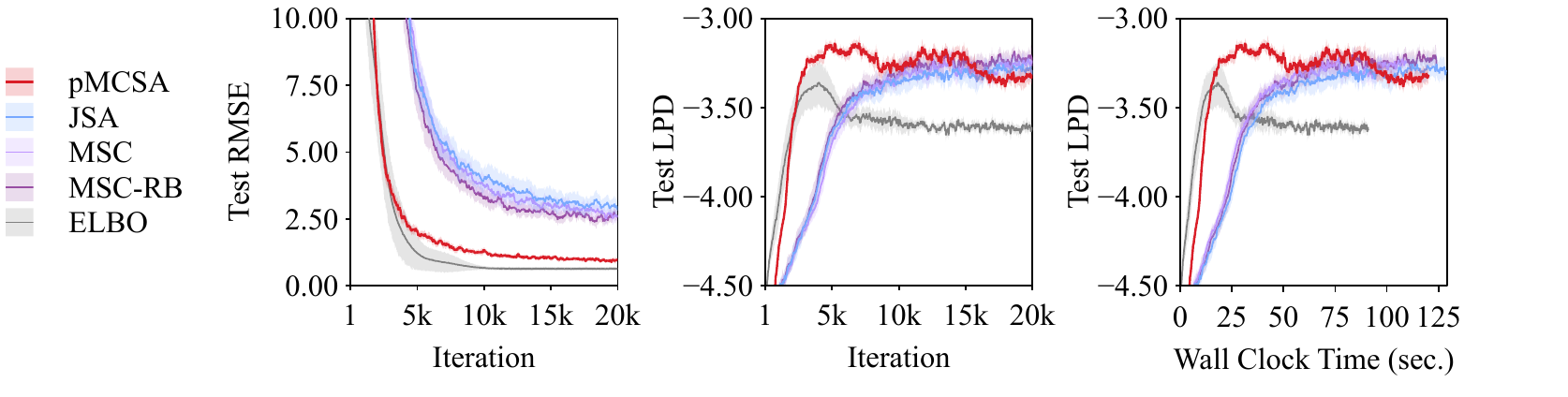}
    \vspace{-0.1in}
  }\vspace{0.05in}\\
  \subfloat[\textsf{airfoil}]{
    \includegraphics[scale=0.9]{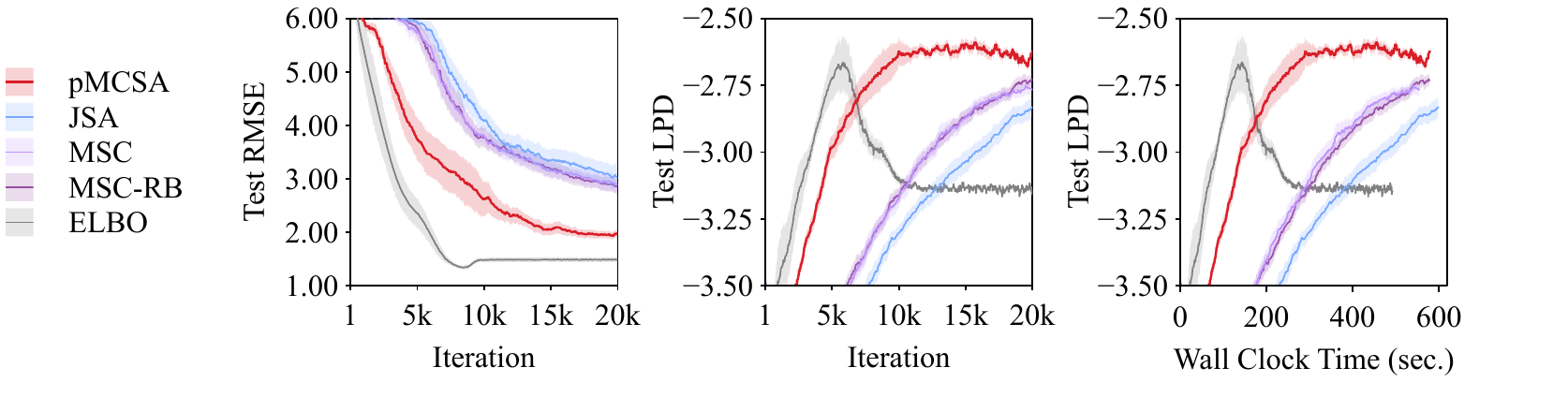}
    \vspace{-0.1in}
  }\vspace{0.05in}\\
  \subfloat[\textsf{boston}]{
    \includegraphics[scale=0.9]{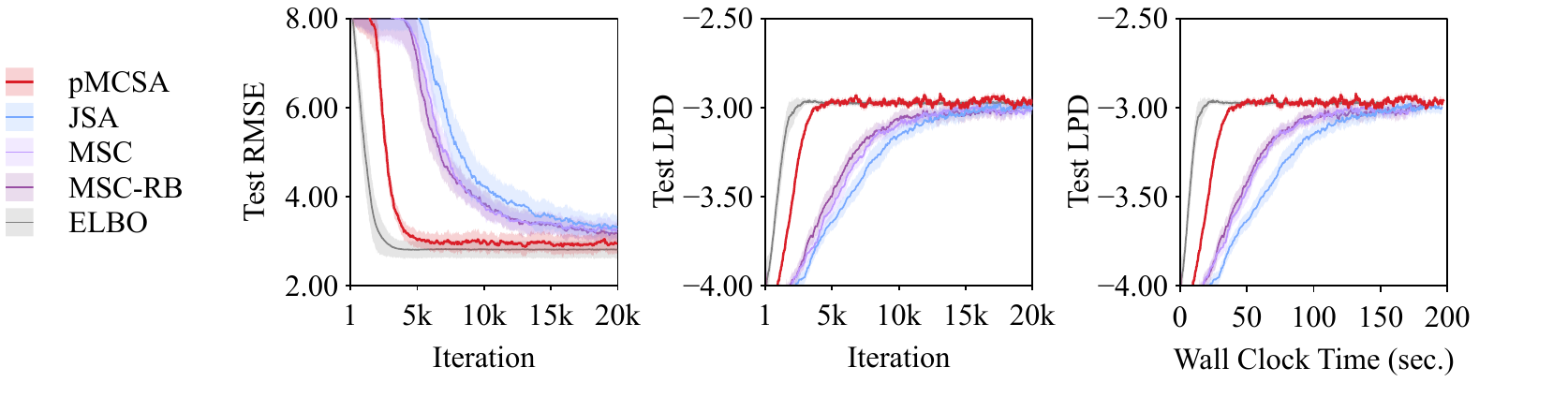}
    \vspace{-0.1in}
  }\vspace{0.05in}\\
  \caption{\textbf{Test root-mean-square error (RMSE) and test log predictive density (LPD) on robust Gaussian process regression.}
  The error bands shows the 95\% bootstrap confidence interval obtained from 20 repetitions.}
\end{figure}
\newpage
\begin{figure}[ht]
  \subfloat[\textsf{energy}]{
    \includegraphics[scale=0.9]{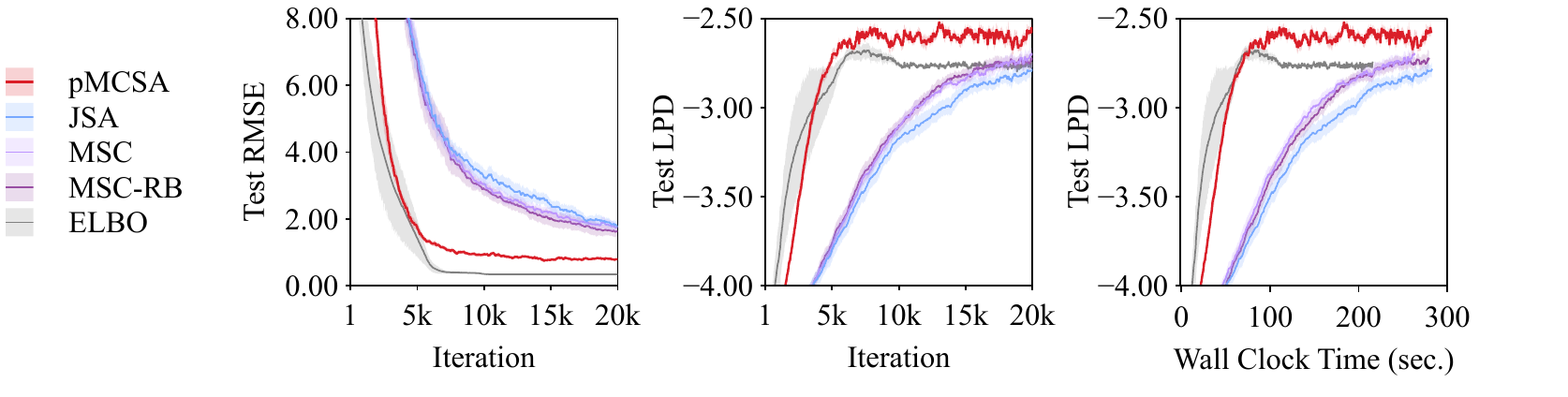}
    \vspace{-0.1in}
  }\vspace{0.05in}\\ 
  \subfloat[\textsf{concrete}]{
    \includegraphics[scale=0.9]{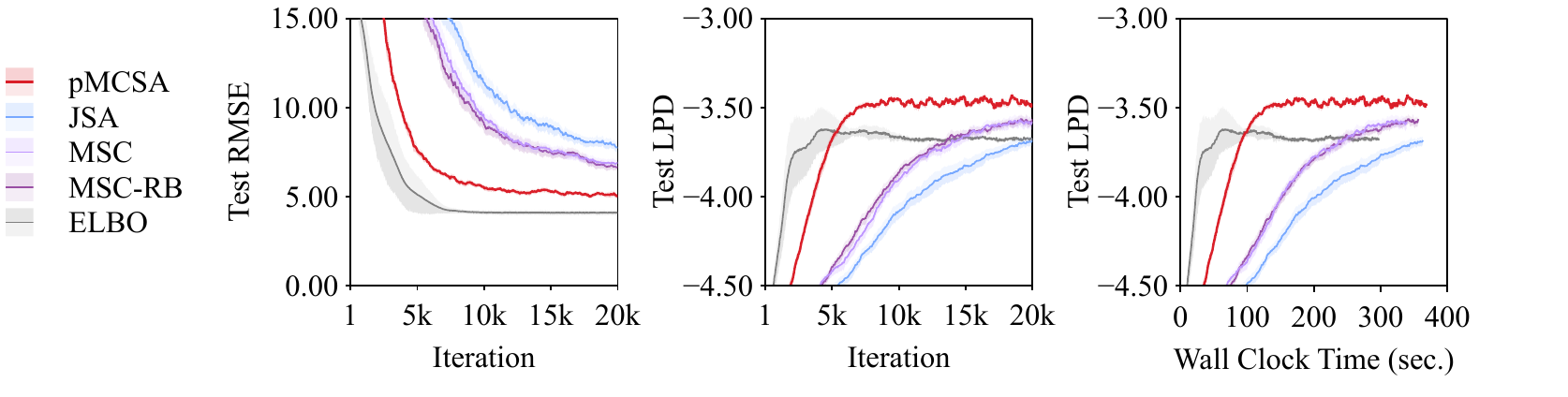}
    \vspace{-0.1in}
  }\vspace{0.05in}\\
  \subfloat[\textsf{wine}]{
    \includegraphics[scale=0.9]{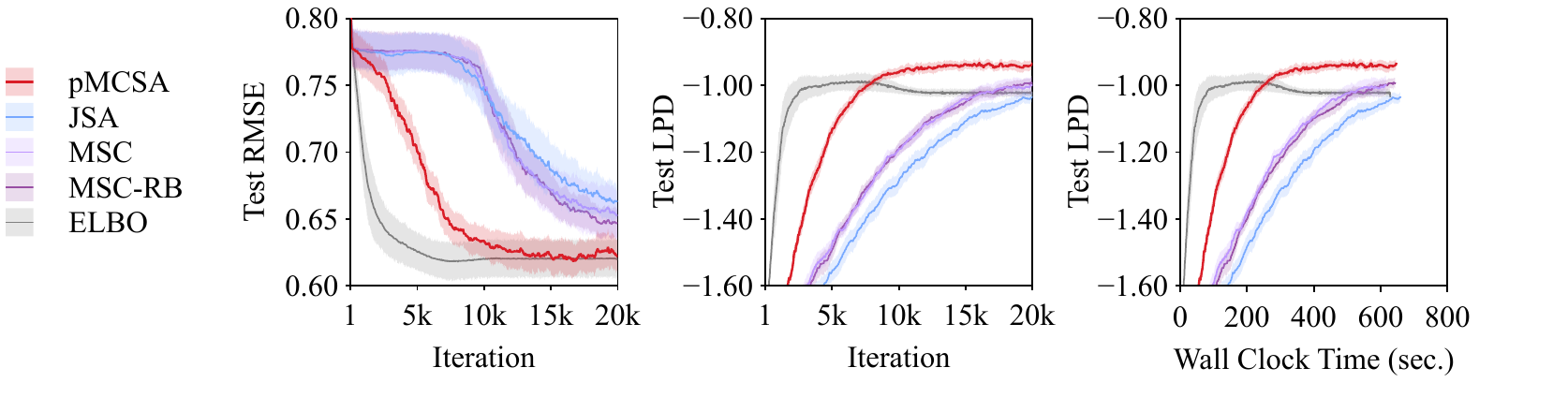}
    \vspace{-0.1in}
  } \vspace{0.05in}\\
  \subfloat[\textsf{gas}]{
    \includegraphics[scale=0.9]{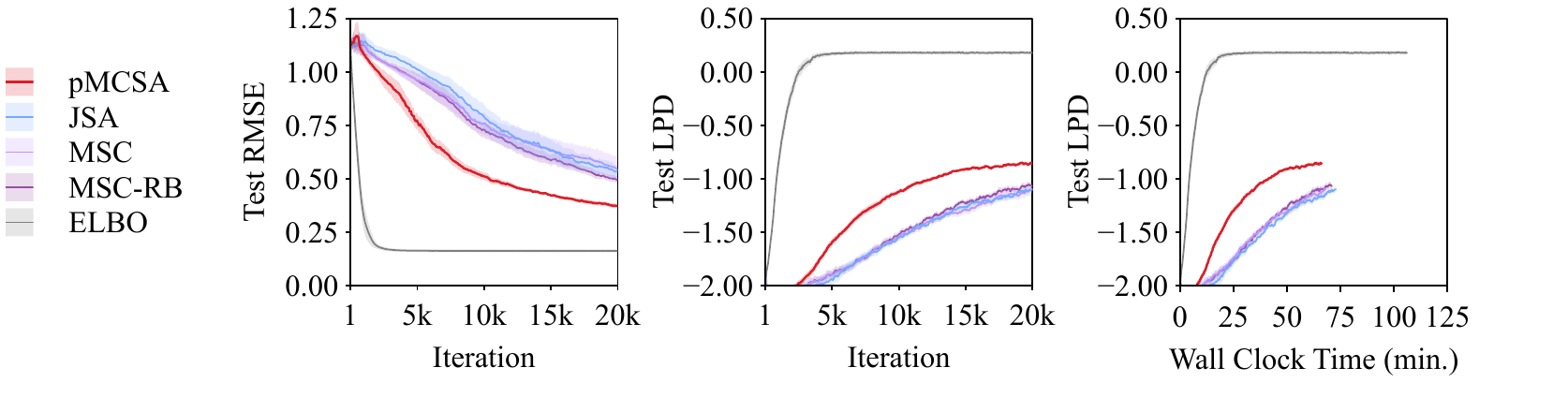}
    \vspace{-0.1in}
  }
  \caption{\textbf{(continued) Test root-mean-square error (RMSE) and test log predictive density (LPD) on robust Gaussian process regression.}
  The error bands show the 95\% bootstrap confidence intervals obtained from 20 independent 90\% train-test splits.
  }
\end{figure}

%%% Local Variables:
%%% TeX-master: "master"
%%% End:

\end{document}